\documentclass[preprint,a4paper,UKenglish]{tgdk}
\usepackage{graphicx}
\usepackage{xcolor}
\usepackage{dsfont}
\usepackage{paralist}
\usepackage[square,numbers]{natbib}
\usepackage[capitalise,nameinlink]{cleveref}
\usepackage{marvosym}
\usepackage{multirow,booktabs}
\usepackage{color}
\usepackage{adjustbox}
\usepackage{makecell}
\usepackage{tabulary}
\usepackage{colortbl}
\usepackage{wrapfig}

\Volume{0}
\Issue{0}
\Article{0}

\newboolean{showcomments}
\setboolean{showcomments}{true} % toggle to show or hide comments
\ifthenelse{\boolean{showcomments}}
  {\newcommand{\nb}[2]{
    \fcolorbox{gray}{yellow}{\bfseries\sffamily\scriptsize#1}
    {$\blacktriangleright$#2$\blacktriangleleft$}
   }
   
  }
  {\newcommand{\nb}[2]{}
   
  }

\newcommand{\added}[1]{\textcolor{black}{#1}}
\newcommand\footnoteref[1]{\protected@xdef\@thefnmark{\ref{#1}}\@footnotemark}

\title{Trust, Accountability, and Autonomy in Knowledge Graph-based AI for Self-determination} 

\titlerunning{Trust, Accountability, and Autonomy in Knowledge Graph-based AI for Self-determination}%optional, please use if title is longer than one line

\author{Luis-Daniel Ib\'{a}\~nez }{Department of Electronics and Computer Science, University of Southampton, UK}{l.d.ibanez@southampton.ac.uk}{0000-0001-6993-0001}{Partially funded by the European Union's Horizon Research and Innovation Actions under Grant Agreement nº 101093216 (UPCAST)}

\author{John Domingue}{Knowledge Media Institute, The Open University, UK}{john.domingue@open.ac.uk}{0000-0001-8439-0293}{}

\author{Sabrina Kirrane}{Institute for Information Systems \& New Media, Vienna University of Economics and Business, Austria}{sabrina.kirrane@wu.ac.at}{0000-0002-6955-7718}{Is currently funded by the FWF Austrian Science Fund and the Internet Foundation Austria under the FWF Elise Richter and netidee SCIENCE programmes as project number V 759-N.}

\author{Oshani Seneviratne}{Department of Computer Science, Rensselaer Polytechnic Institute, USA}{senevo@rpi.edu}{0000-0001-8518-917X}{Partially funded by NSF IUCRC CRAFT center research grant (CRAFT Grant \#22008) and the Algorand Centres of Excellence programme managed by Algorand Foundation. The opinions expressed in this publication do not necessarily represent the views of NSF IUCRC CRAFT or the Algorand Foundation.}

\author{Aisling Third}{Knowledge Media Institute, The Open University, UK}{aisling.third@open.ac.uk}{XXXX-XXXX-XXXX-XXXX}{}

\author{Maria-Esther Vidal}{Leibniz University of Hannover \& TIB-Leibniz Information Centre of Science and Technology, Germany}{maria.vidal@tib.eu}{0000-0003-1160-8727}{Partially funded by Leibniz Association, program ``Leibniz Best Minds: Programme for Women Professors'', project TrustKG-Transforming Data in Trustable Insights; Grant P99/2020}

\authorrunning{Ib\'{a}\~nez et al.}

\Copyright{Ib\'{a}\~nez, Domingue, Kirrane, Seneviratne, Third, and Vidal}

\begin{document}
\nolinenumbers
\maketitle

\begin{abstract}

Knowledge Graphs (KGs) have emerged as fundamental platforms for powering intelligent decision-making and a wide range of Artificial Intelligence (AI) services across major corporations such as Google, Walmart, and AirBnb. KGs complement Machine Learning (ML) algorithms by providing data context and semantics, thereby enabling further inference and question-answering capabilities. The integration of KGs with neuronal learning (e.g., Large Language Models (LLMs)) is currently a topic of active research, commonly named neuro-symbolic AI. 
Despite the numerous benefits that can be accomplished with KG-based AI, its growing ubiquity within online services may result in the loss of self-determination for citizens as a fundamental societal issue. The more we rely on these technologies, which are often centralised, the less citizens will be able to determine their own destinies. To counter this threat, AI regulation, such as the European Union (EU) AI Act, is being proposed in certain regions. The regulation sets what technologists need to do, leading to questions concerning: How can the output of AI systems be trusted? What is needed to ensure that the data fuelling and the inner workings of these artefacts are transparent? How can AI be made accountable for its decision-making?
This paper conceptualises the foundational topics and research pillars to support KG-based AI for self-determination. Drawing upon this conceptual framework, challenges and opportunities for citizen self-determination are illustrated and analysed in a real-world scenario. As a result, we propose a research agenda aimed at accomplishing the recommended objectives.
%categorised in short-, middle- and long-term objectives.  

%The seminar will explore these questions from a KG-based AI viewpoint. In particular, the seminar will comprise three pillars - trust, accountability, and self-determination - which will form the structure for the first part of the event. The latter parts of the seminar will allow for the free-forming of groups which will be tied together on the final day.

\end{abstract}

\section{Introduction}
\label{sec:introduction}

Modern \emph{Artificial Intelligence} (AI) can be traced back to a workshop held at Dartmouth College in the summer of 1956~\cite{luger_modern_2021} \added{ and is most commonly defined as the use of computers to simulate human intelligence, in particular human reasoning, learning, and problem-solving. Since 1956,} AI has lived through times of increased interest and funding, and also `AI Winters', such as after the 1974 Lighthill report~\cite{lighthill_artifical_1974}, when overall funding was reduced. 
Over the last few years, however, funding and interest in AI have been high and exploded in November 2022, when ChatGPT, a type of Generative AI, was announced by OpenAI, exposing the power of Large Language Models (LLMs) to the general public. Since its release, ChatGPT has become the fastest-growing app in history, reaching 100M users in just two months, and is now estimated to have 200M users. Generative AI will continue to grow following a significant investment by Microsoft into OpenAI and announcements by Microsoft and Google on how Generative AI will be embedded in future products~\cite{gordon_chatgpt_2023}.
\added{Data-centric AI~\cite{whang2023data} recognises the immense value of data as crucial resources for training, optimising, and evaluating AI systems. Databricks, a prominent AI company, has defined data-centric AI as the challenge of designing processes for data collection, labelling, and quality monitoring in machine learning (ML) datasets~\cite{polyzotis_what_2021} highlighting the need for continuous re-running and re-training, actionable monitoring, and the difficulties of incorporating data inaccessible to human annotators due to privacy concerns as primary research directions. Knowledge Graphs have been used both as a resource and as a structure to support data-centric AI processes.}
%The reliance on high-quality, relevant, and diverse data is essential in this approach, as it drives the algorithms and models used in AI. 
%
The term \emph{Knowledge Graph} (KG) was first introduced by Google in 2012, and \added{is usually defined as a type of knowledge structure that uses a graph data model to integrate data. KGs are} strongly linked to the work of the Semantic Web community, which first began in around 2001 and was introduced in a seminal paper by Tim Berners-Lee \cite{berners-lee_semantic_2001}. \added{The Semantic Web initiative produced a stack of web standards on which KGs are based. These include the Resource Description Framework (RDF), where data is encoded as subject-predicate-object triples, and the Web Ontology Language (OWL), a set of web-based languages mostly based on description logic. The common theme of these semantic representations is that they facilitate the publishing, use, and re-use of data at the web scale. In particular, they allow disparate heterogeneous data sources to be integrated continuously at scale.}
Over the past decade, \added{KGs have} become a mainstay for a number of key large-scale applications found online. For example, KGs underpin Google Search, which saw 5,900,000 searches in just one minute in April 2022. Similarly, the same minute saw 1,700,000 pieces of content shared on Facebook, 1,000,000 hours streamed, and 347,200 tweets shared on Twitter. All of this content and data are linked to a plethora of AI services that have increasingly been based on KGs, \added{as mentioned above, }founded upon machine-readable data and schema representations based on a web stack of standards. AI services cover a wide number of areas, including content recommendation, user input prediction, as well as large-scale search and discovery and form the basis for the business models of companies like Google, Netflix, Spotify, and Facebook. 
\added{Given the above we define KG-based AI as an AI system (replicating some aspect of human intelligence) based on a KG possibly using the web standards produced by the Semantic Web community.}

In addition to privacy concerns, there has been a growing worry about how personal data can be abused and, thus, how AI services impinge on citizen rights. For example, the over-centralisation of data and its misuse led Sir Tim Berners-Lee to call the Web ‘anti-human’ in an interview in 2018~\cite{brooker_i_2018}. Since 2016, hundreds of United States (US) Immigration and Customs Enforcement employees have faced investigations into abuse of confidential law enforcement databases, including stalking and harassment, to passing data to criminals~\cite{mehrotra2023ice}.
The subject of much of the proposed legislation today is ensuring that digital platforms, including AI platforms, provide real societal benefit. Within Europe, the proposed  European Union (EU) AI Act\footnote{\url{https://artificialintelligenceact.eu/}} aims to support safe AI that respects fundamental human rights. The regulation sets what technologists need to do.  The concept of data \emph{self-determination}, which is often used in a legal context, implies that individuals are not only aware of who knows what about them but can also influence data processing that concerns them \cite{kukutai_pushing_2021}. Given that nowadays, data processing is conducted by opaque AI algorithms behind corporate firewalls, sometimes even without our knowledge, data self-determination is harder than ever before. When it comes to trust in web data and services, Berners-Lee and Fischetti \cite{berners-lee_weaving_2000} envisaged an “Oh yeah?” button embedded into Web browsers that would provide justifications as to why a page or a service should be trusted. Alas, their vision was never realised in popular web browsers\footnote{However, a linked browser prototype, the Tabulator, incorporated this feature in an \emph{Justification UI} (\url{http://dig.csail.mit.edu/TAMI/2008/JustificationUI/howto.html\#useTab}).}. Instead, we have dedicated websites, e.g., the Ecommerce Europe Trustmark\footnote{\url{ https://ecommercetrustmark.eu/}} that are used to perform company reputation checks and fact-checking websites, such as Snopes\footnote{\url{https://www.snopes.com/}}, that can be used to check the validity of information posted online. Although some automated fact-checking techniques have been proposed \cite{qudus_hybridfc_2022}, they are used solely for developing trust in information resources and cannot provide any guarantees with respect to AI-based data processing. Moving beyond trust towards accountability, policies have already been used to specify legal data processing requirements that serve as the basis for automated compliance checking, for example, \cite{ko_pronto_2018}. But what happens when service providers or AI algorithms do not comply? How far can technology go in terms of helping us to determine non-compliance and to make service providers accountable for their actions?  

\begin{figure}[t!]
    \centering
    \includegraphics[width=\textwidth]{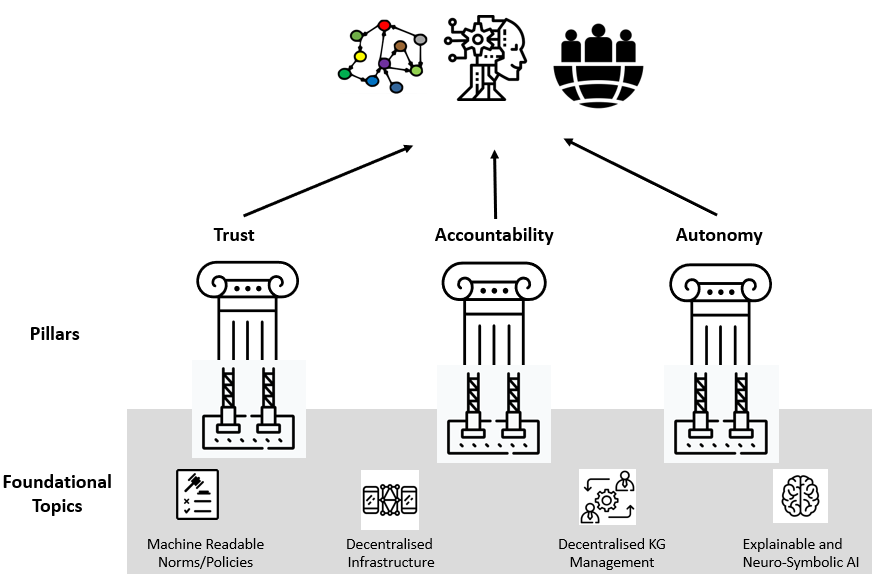}
    \caption{\textbf{KG-based AI for Self-determination Conceptualisation.} KG-based AI for self-determination is supported by the pillars of trust, accountability, and autonomy, built upon the foundational topics of machine-readable norms and policies; decentralised infrastructure; decentralised KG management; and explainable neuro-symbolic AI.}
    \label{fig:pillars}
\end{figure}
\added{In this paper, we propose a research agenda for ensuring that KG-based AI approaches contribute to user self-determination instead of hindering it. Our vision, which is depicted in Figure~\ref{fig:pillars}, is structured around three pillar research topics - trust, accountability, and autonomy - that represent the desired goals for how AI can benefit society and facilitate self-determination. The pillars combine fundamental principles of the proposed EU AI Act and self-determination theory. Both trust and accountability are imperative for safeguarding against adverse impact caused by AI systems, while autonomy is critical for ensuring individuals are able to determine their own destiny. The pillars are supported via four foundational research topics - machine-readable norms and policies are needed for humans to declare regulatory frameworks, privacy and usage constraints that can be interpreted by the machines that process their data; explainable neuro-symbolic AI to clearly communicate and prove the decisions AI systems make; and decentralised KG management and decentralised infrastructure to provide alternatives to approaches where a central entity controls a whole process, that are prone to abuse of power}. We posit the following research questions:

\begin{description}    
    %\item[Q1] What is required to ensure that the data fueling and the inner workings of AI artefacts are transparent? 
    \item[Q1] What are the key requirements for an AI system to produce trustable results?
    \item[Q2] How can AI be made accountable for its decision-making?
    \item[Q3] How can citizens maintain autonomy as users or subjects of KG-based AI systems?
\end{description}

\begin{figure}[t!]
    \centering
    \includegraphics[width=\textwidth]{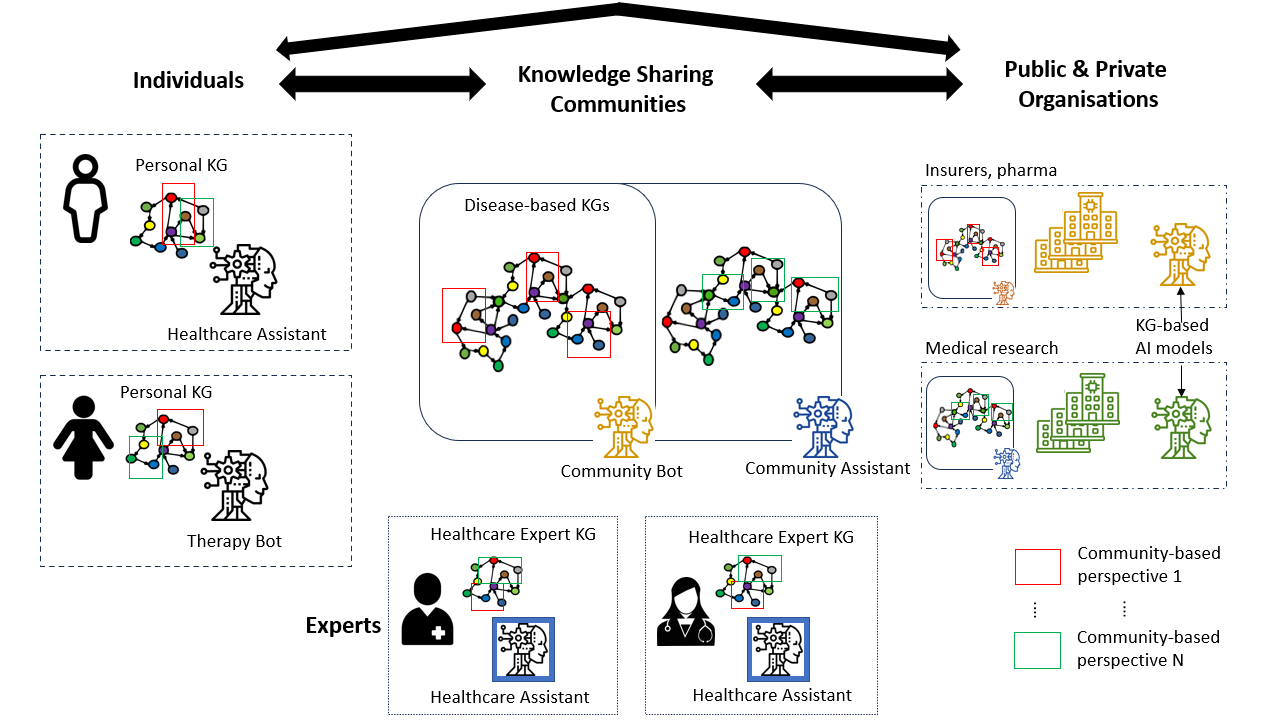}
    \caption{\textbf{Illustrative Scenario for KG-Based AIs in the healthcare domain.} Individuals use AI assistants to make sense of data collected in their PKGs. They may also share perspectives of their PKGs with other individuals and healthcare experts in knowledge-sharing communities that aggregate and curate data to power AI services for the benefit of all members. Public and private organisations can negotiate access to data from communities and individuals to train KG-based AI models, which in turn are used to build services for them.}
    \label{fig:scenario}
\end{figure}

\noindent \added{In order to facilitate exposition, we ground our discussion in a healthcare scenario inspired by the recently proposed regulation on European Health Data Space\footnote{https://eur-lex.europa.eu/legal-content/EN/TXT/?uri=celex\%3A52022PC0197} that aims to ensure that \emph{"natural persons in the EU have increased control in practise over their electronic health data"} and to facilitate access to health data by various stakeholders in order to \emph{"promote better diagnosis, treatment and well-being of natural persons, and lead to better and well-informed policies"}. The proposed healthcare scenario, which is illustrated in Figure~\ref{fig:scenario}, is comprised of the following actors and interactions:}

\begin{description}  
\item[Individuals] manage their Personal Knowledge Graphs (PKGs) (aligned with the original Semantic Web vision and  modern interpretations~\cite{balog_personal_2019,ilkou_personal_2022}). They collect knowledge about their medical conditions, symptoms, treatments, reactions to treatments, etc. Individuals get services from KG-based AI applications that utilise their PKGs, e.g., therapy bots or health assistants. 

\item[Experts] in healthcare also have PKGs where they collect their knowledge about diseases, results of the treatments they have suggested in the past, links to general medical knowledge graphs, etc. Experts may also be assisted by KG-based AI models.       

\item[Knowledge sharing communities] are spaces where individuals and healthcare experts may share subsets of their PKGs in the context of specific knowledge, e.g., diseases. We call these \emph{community-based perspectives}. Perspectives from different contributors are aggregated into community KGs (e.g., disease-based). AI applications use these KGs for community benefit, e.g., assessing if a treatment that worked for an individual may work on a different one.

\item[Public and private organisations] may negotiate access to data and knowledge from communities to train large KG-based AI models to either improve internal processes or power products sold to communities, experts, or individuals, completing the cycle.
\end{description}

 %Individuals may organise or join communities to share data from their PKGs, e.g., their treatments and how they have worked for them, with applications offering value on top of the shared knowledge. Healthcare professionals such as doctors may also use PKGs to collect their experience treating certain diseases and AI models to assist them. Doctors may also participate in communities around a certain disease. 
 
%\end{minipage}

\smallskip
The remainder of the paper is structured as follows: \added{\Cref{sec:background} introduces the necessary background in terms of KG-based AI.} \Cref{sec:pillars} highlights the importance of trust, accountability, and autonomy when it comes to ensuring that AI benefits society. \Cref{sec:toolbox} presents several KG based tools and techniques that can be used to facilitate  trust, accountability, and self-determination.  In \Cref{sec:roadmap}, we propose a research roadmap that includes several challenges and opportunities for KG-based AI that benefits individuals and society. Finally, we conclude and outline important first steps in \Cref{sec:conclusion}.

\added{\section{Knowledge Graph-based AI} \label{sec:background}}

\noindent\added{In his seminal publication, \emph{"Thinking, Fast and Slow"}, Daniel Kahneman \cite{kahneman2011thinking} presents a comprehensive theory of human intelligence, offering profound insights into the workings of the human mind. This groundbreaking work separates intuition from rationality when approaching problem-solving tasks, defining them as two sets of abilities or \emph{systems}. System 1 operates at an unconscious level, generating responses effortlessly and swiftly. In contrast, System 2 requires conscious attention and concentration, enabling the generation of responses needing complex computations. Kahneman's characterisation of mental cognition aligns with statistical and symbolic learning models that seek to simulate human thinking processes~\cite{DBLP:conf/aaai/BoochFHKLLLMMRS21}. These systems are known as neuro-symbolic systems \cite{10.1145/3586163}, and there is a growing interest in emerging hybrid approaches that aim to integrate cognitive capabilities. Specifically, they strive to combine the power of neural networks, such as LLMs, with the interpretability offered by symbolic processing, particularly semantic reasoning over KGs.}

\added{\subsection{Knowledge Graphs}}
\noindent\added{Google first introduced KGs in 2012 when they enabled `Knowledge Panels' containing descriptions including pictures for search items. For example, if one types in `London' to Google Search, the Knowledge Panel displays pictures, the current weather, a map, directions, elevation and related entities (e.g. Paris). The seed for the Google KG was Freebase - a community knowledge base initially launched in 2007 with an add-on RDF service launched at the International Semantic Web Conference in 2008. In 2010, Google bought Metaweb, the company that owned Freebase and extended the knowledge base into the Google KG\footnote{\url{https://en.wikipedia.org/wiki/Schema.org}}.}

\added{In 2011, Bing, Google and Yahoo! launched Schema.org, a reference website for common data schemas related to web search. The proposal was that website owners would use the published schemas alongside Semantic Web standards such as RDFa and JSON-LD. A number of the schemas, such as Organisation, influence the results returned by Google KG search. Schema.org is an example of a shared vocabulary for semantic representation; the use of such vocabularies or ontologies in KGs, along with the ability to map between equivalent schemas in them, enables integration of heterogeneous data at scale.}

\added{Today, KGs are used in a wide range of areas and products outside of search. For example, Netflix, Amazon, and Facebook all use KGs as the foundation for their recommendation engines for television programmes and films, consumer products and posts\footnote{\url{https://builtin.com/data-science/knowledge-graph}}, whereas in the healthcare sector, KGs are used to integrate medical knowledge and support drug discovery.\footnote{\url{https://www.wisecube.ai/blog/20-real-world-industrial-applications-of-knowledge-graphs/}}}

\added{\subsection{Large Language Models}}
\added{A Large Language Model (LLM) is a specialized machine learning model constructed using a transformer architecture, a category of deep neural networks~\cite{DBLP:journals/corr/abs-2303-18223}. LLMs are primarily designed for predicting the next word in a sequence, making them flexible tools for various text processing tasks, such as text generation, summarization, translation, and text completion. Examples of existing LLMs include OpenAI's ChatGPT~\cite{schulman2022chatgpt} and Google's PALM~\cite{chowdhery2022palm}. These models have demonstrated high performance in Natural Language Processing (NLP) tasks like code generation, text generation, tool manipulation, and comprehension across diverse domains, often achieving high-quality results in zero-shot and few-shot settings. This success has stimulated advancements in LLM architectures, training techniques, prompt engineering, and question answering~\cite{DBLP:journals/corr/abs-2210-13966}.}

\added{Despite their unquestionable capabilities in emulating human-like conversations, there is an ongoing debate regarding the intelligence exhibited by LLMs, particularly, since their fluency in language does not necessarily imply a cognitive understanding of real-world problems~\cite{DBLP:journals/corr/abs-2210-13966}. Additionally, LLMs can only learn knowledge when it appears in the training data and may perform badly when answering questions involving long-tailed facts ~\cite{DBLP:journals/corr/abs-2308-14217}. Moreover, they may struggle to absorb new knowledge and are not easy to audit~\cite{DBLP:journals/corr/abs-2302-08500}, suggesting potential risks of discrimination and information hazards.}

\added{\subsection{Neurosymbolic AI}}
\added{LLMs-- and machine learning models in general-- are trained on extensive datasets, resulting in high-quality outcomes whenever applied to specific prediction tasks. However, LLMs-- like OpenAI's ChatGPT~\cite{schulman2022chatgpt}-- lack of causal understanding and may hallucinate in cases which are not statistical in nature (e.g., memories or explanations)~\cite{DBLP:conf/nesy/HammondL23}. On the other hand, Symbolic AI systems are capable of emulating human-like conscious processes required for causality, logic and counterfactual reasoning, and maintaining long-term memory. As a result, symbolic systems can empower LLMs by modelling human learning and combining knowledge extracted (e.g., from KGs) to formulate prompts that allow for a more fluent communication with users.  
}

\added{Neuro-symbolic AI provides the basis for integrating the discrete approaches implemented by Symbolic AI with high-dimensional vector spaces managed by LLMs. They must decide when and how to combine both systems, e.g., following a principled integration (combining neural and symbolic while maintaining a clear separation between their roles and representations) or integrated (e.g., a symbolic reasoner integrated into the tuning process of an LLM).
Recently, van Bekkum et al.~\cite{DBLP:journals/apin/BekkumBHMT21} propose 17 fundamental design patterns to model neuro-symbolic systems. These patterns encompass many scenarios where the symbiotic relationship between symbolic reasoning and ML models becomes apparent. Since these combinations may enable symbolic reasoning and enhance contextual knowledge, neuro-symbolic systems may empower explainability and, as a result, also improve transparency by showing how a system works based on the symbolic explanations deduced by the hybrid system. 
}
\bigskip
\section{KG-based AI that Benefits Individuals and Society}
\label{sec:pillars}
\added{Considering our vision that KG-based AI can facilitate self-determination}, we start by discussing the pertinent role played by trust, accountability, and autonomy when it comes to ensuring that AI benefits society. In each case, we highlight existing challenges and present arguments in favour of KG-based AI system. 

\subsection{Trust and KG-based AI}
\added{One of the primary objectives of the proposed EU AI Act is the \emph{"development of an ecosystem of trust by proposing a legal framework for trustworthy AI"}}. The Merriam-Webster dictionary definition of trust includes a \emph{“firm belief in the reliability, truth, or ability of someone or something”}~\cite{trust-definition}. 
Questions we address in this paper include understanding how KG-based AI systems can demonstrate reliability, truth, and ability through mechanisms, which add transparency to all elements involved in KG-reasoning. These include: comprehensive provenance tracking of data sources and data elements used for any output; understanding repeatability for all KG-based AI reasoning (e.g., if datasets are altered or disappear altogether, or if other reasoning methods, such as LLMs, are involved); and alleviation mechanisms when KG-based AI system responses are untruthful. %Distributed ledgers are decentralised trust mechanisms where trust is provided by a combination of encryption, comprehensive data sharing, and community consensus mechanisms for processing transactional data. 
% What can the KG community learn from distributed ledgers on trust?

The proliferation of misinformation on the internet has risen significantly in recent years, coinciding with the advancements in generative AI technologies. As AI becomes more sophisticated, it has inadvertently provided tools and techniques for the creation and dissemination of false information, leading to widespread confusion and societal harm~\cite{ciampaglia2018research,zhou2023synthetic}. 
For instance, AI-generated deepfake videos have become a concerning source of misinformation. Deepfakes use AI algorithms to manipulate and superimpose faces onto existing videos, making it difficult to discern real from fabricated content~\cite{westerlund2019emergence}. This technology has been used to create fake videos of public figures saying or doing things they never actually did, leading to potential defamation and manipulation of public opinion.
AI-powered chatbots and automated accounts on social media platforms have been employed to spread false information and manipulate public sentiment. These bots can mimic human-like conversations and flood social media platforms with fake news, propaganda, and divisive narratives, influencing public opinion and sowing discord, and have even contributed to misinformation in medical literature~\cite{lubowitz2023chatgpt}.
AI-powered recommendation algorithms used by platforms like social media and video-sharing websites can inadvertently contribute to the spread of misinformation. These algorithms aim to maximise user engagement by suggesting content based on user preferences and behaviour. They can create filter bubbles, reinforcing users' existing beliefs and exposing them to a limited range of perspectives, potentially amplifying false information and preventing users from accessing accurate and diverse sources of information~\cite{pariser2011filter}.

Amidst these challenges, KG technologies have emerged as a potential solution to curb misinformation and enhance trust. Leveraging the power of crowd-supplied and verified knowledge sources, such as Wikidata~\cite{vrandevcic2014wikidata}, KGs enable comprehensive fact-checking capabilities. By integrating diverse and reliable information from various trusted sources, these graphs can potentially identify and flag misleading or inaccurate content more effectively. By utilising the collective intelligence of a crowd, KG technologies empower users to contribute to the verification process, enhancing the accuracy and credibility of the information presented. Through collaborative efforts and the utilisation of KG technologies, it is possible to combat the rising tide of misinformation, safeguarding the integrity of online information and fostering a more informed digital society. Coupled with distributed ledgers, it has been proposed that KG-based AI can combat misinformation on the web~\cite{seneviratne2022blockchain}. There is already a growing body of work in this space, which shows some promise. For example, Mayank et al.~\cite{mayank2022deap} and Koloski et al.~\cite{koloski2022knowledge} describe systems that leverage KGs to detect fake news; Kou et al.~\cite{kou2022hc} and and Shang et al.~\cite{shang2022privacy} describe how crowd-sourced KGs can be used to mitigate COVID-19 misinformation; Kazenoff et al.~\cite{kazenoff2020semantic} use semantic graph analysis to detect cryptocurrency scams propogating in social media. 

\subsection{Accountability and KG-based AI}
\added{According to the proposed EU AI Act, when it comes to high-risk AI, \emph{"accuracy, reliability and transparency is particularly important to avoid adverse impacts, retain public trust and ensure accountability and effective redress"}}. Accountability in a KG-based AI context assumes that data scientists, computer scientists, and software engineers will follow best practices and ensure compliance with relevant legislation.  In the purely symbolic world, such properties can be achieved via consistency and compliance checking based on formal requirements specified in policy languages such as LegalRuleML~\cite{Athan2015} and ODRL~\cite{ianella_odrl_2018}. When it comes to the sub-symbolic world, these principles are particularly challenging, as ML algorithms are often opaque and could potentially infer confidential information during the training process. In recent years, various Explainable AI (XAI) techniques have been used to build or applied to the output of models such that they can be interpreted and understood by various stakeholders~\cite{pmlr-v70-koh17a}. In the context of KG-based AI this will require the intersection between two strains of explainability: the explanation of why a statement is in the KG that supports the AI, and the explanation of how the model used the statements from the KG to reach a particular decision. KGs can also be used to support the modelling, capturing, and auditing of records useful for accountability throughout the system life cycle~\cite{naja_accountability_2021}

When it comes to AI and accountability technical research should go hand in hand with the interdisciplinary research conducted in communities like FaccT\footnote{\url{https://facctconference.org/index.html}}. A recent paper \cite{cooper_accountability_2022} revisited the four barriers of accountability that were developed in the 1990s for accountability of computerised systems in the light of the rise of AI, finding that they are even more important than before. The main barrier is the problem of \emph{many hands} - the large amount of actors involved in the construction of an AI service creates difficulties in the assignment of responsibilities in case of harm. Advancing efficient provenance collection, and verifiability will be the key technical intervention to overcome this barrier. Fields such as data science require strong guarantees for provenance to build context-aware KGs~\cite{seneviratne2020data}. 
Similar to explainability, we consider two different approaches zhat need to be combined:  the provenance of statements in the KG and the provenance of the pipeline that was followed to construct the ML model. %The latter will require the   

%In the seminar, we will investigate how KGs can be used to enhance the accuracy, reliability, and transparency of AI.

\subsection{Autonomy and KG-based AI}
\added{Alongside accountability and trust, the third pillar needed to support self-determination is \emph{autonomy}, defined from a self-determination theory\footnote{\url{https://en.wikipedia.org/wiki/Self-determination\_theory}} perspective as \emph{"the belief that one can choose their own behaviors and actions"}}. In the current context, we take this to mean that individuals should be able to make their own decisions about their uses of KG-based AI and about its uses of their data (and have their wishes respected). Assuming that AI systems can be made to be trustable and accountable, how can we best support autonomy in this way? That is to say, if we can know that an AI will behave in a desired and known way, and that its decisions and processes are transparent and traceable, how can we express and enable control over what it does in regard to an individual? A number of approaches have emerged in recent years which facilitate individuals' data sovereignty and how they represent and express their identity online. 

The concept of a PKG--  introduced in  our illustrative scenario-- is one means of facilitating autonomy; Solid pods~\cite{sambra2016solid,mansour2016demonstration} are secure decentralised data stores accessible through standard semantic interfaces for applications that generate and consume linked data. Currently, the default model on the Web is for service providers to host and control access to user data by means of a user account. This denies autonomy to the individuals concerned since all access is mediated via applications and interfaces designed and controlled by service providers. The PKG model is that personal data is independent of any application; PKGs are the primary source of data under the control of individuals, and they mediate service access via standard interfaces. On top of shifting control away from service providers, this approach makes it technically simpler to implement data usage policies, as they can be stored with the data and evaluated at the PKG level. 

One prominent way of achieving the second goal is through the notion of Self-Sovereign Identity (SSI) \cite{der2017selfsovereign}. Traditional digital identity (e.g., as in OpenID Authentication \cite{rfc6749}) has been modelled in terms of Identity Providers (IdPs). An individual and an IdP establish a relationship, and the IdP generates a digital identity for them. If the individual wants to authenticate with a third party, the IdP confirms the relationship to them and then asserts that identity to the relying party. Crucially, sovereignty over that identity and decisions about who can see it, the data associated with it, or whether it continues to exist are taken by the IdP. With SSI, an individual generates their \emph{own} digital identity (e.g., a cryptographic key pair), makes their own identity assertions, and therefore has full control over that identity, with correlations between two identities (digital or physical) relying explicitly on attestation by others, and trust relationships with them\footnote{As it ultimately does in traditional digital identity, where trust in a small number of well-known IdPs serves as a simplified proxy for more detailed or fine-grained considerations of trust networks.}. The autonomy enabled by SSI makes \emph{selective disclosure} possible, meaning that what identity information gets shared with whom can be made contextually and on a case-by-base basis - much like presenting different aspects of ones personal identity in daily life (e.g., work and home personas). 

Considerations of identity pervade any technical considerations for safeguarding self-determination. It seems uncontroversial that there will be scenarios in which an individual’s identity is relevant to what they wish to do with a KG-based AI, whether in training, KG contents, or inference, and indeed, even where anonymity is desired, identity must be considered in order to avoid revealing it. Identity is also fundamental to the concept of trust; trust in a person, organisation, system, AI model, KG, etc., is useful only in so far as it is possible to identify relevant entities as needed, and accountability cannot be tracked or apportioned without it.
We consider autonomy in terms of the identity, data, and sovereignty afforded to an individual or organisation in terms of what they or others communicate to a KG-based AI ecosystem or elements thereof, what they or others receive from those, and what happens to those (including respect of choices) as data is processed in the ecosystem, with each of these evaluated through the lenses of selective disclosure, relevant identities, and utility.
%This theme of the seminar will explore how KG-based AI would operate efficiently in the self-determined contexts outlined above. Specifically, how KG data would be managed, how KG reasoning, for example, based on (federated) SPARQL querying, would function. Also, what technical mechanisms and approaches from DIDS, VCs and distributed ledgers could support self-determined KG-based AI.

\section{A KG Toolbox for Trust, Accountability, and Autonomy}
\label{sec:toolbox}
 
\added{In order to ground our pillars, we motivate and introduce relevant literature and highlight open research challenges and opportunities concerning our foundational topics: machine-readable norms and policies; decentralised infrastructure; decentralised KG management; and explainable neuro-symbolic AI, each of which plays a pivotal role in facilitating trust, accountability, and autonomy in KG-based~AI.}

\subsection{Machine-readable Norms and Policies}

When it comes to KG-based AI, norms and policies could potentially be used to inform data processing based on legal requirements, social norms, privacy preferences, and licensing. Legal documents are designed in natural language for human consumption, thus in order to enable machines and automated agents to evaluate and enforce the agreements embodied in documents, we need to translate them to formats they can read and process efficiently.

\bigskip
\noindent
\added{\textbf{Norm and Policy Encoding.}}
Languages to express policies, including but not limited to data access, can be categorised as either general or specific. In the former, the syntax caters to a diverse range of functional requirements (e.g., access control, query answering, service discovery, negotiation), whereas the latter focuses on just one functional requirement. In the early days of the Semantic Web, research into general policy languages that leverage semantic technologies (e.g., KAoS~\cite{uszok2003kaos}, Rei~\cite{kagal2002rei}, AIR~\cite{khandelwal2010analyzing}, and Protune~\cite{bonatti2010protune}) was an active area of research. \added{However, despite the huge potential offered by these general purpose languages to date none of them achieved mainstream adoption \cite{kirrane_access_2016}.}
More recently, researchers have proposed ontologies that can be used to represent licenses, privacy preferences, and regulatory obligations \cite{kirrane_intelligent_2021}. When it comes to the legal domain specifically, Semantic Web researchers have proposed cross-domain ontologies that can be used to encode legal text in a machine-readable format using LegalRuleML\footnote{\url{https://docs.oasis-open.org/legalruleml/legalruleml-core-spec/v1.0/os/legalruleml-core-spec-v1.0-os.html}} and adaptations thereof (e.g.,~\cite{athan2013, palmirani2011}). Others focus on facilitating legal document indexing and search using the European Law Identifier (ELI)\footnote{\url{https://eur-lex.europa.eu/legal-content/EN/TXT/?uri=CELEX:52012XG1026(01)}} and the European Case Law Identifier (ECLI)\footnote{\url{https://eur-lex.europa.eu/legal-content/EN/TXT/?uri=CELEX:52011XG0429(01)}} (e.g.,~\cite{oksanen2019, chalkidis2017}), or bridging the gap between the EU and member state legal terminology (e.g.,~\cite{ajani2016, boella2019}). Besides these cross-domain activities,  there have also been various domain-specific initiatives.  For instance, the ELI ontology has to be extended to facilitate the encoding of the text of the General Data Protection Regulation (GDPR)\footnote{ \url{https://eur-lex.europa.eu/legal-content/EN/TXT/?uri=CELEX\%3A32016R0679&qid=1681238509224}} (e.g.,~\cite{pandit2018}).  While others have focused specifically on modelling privacy policies (e.g.,~\cite{oltramari2018, palmirani2018}).  The \added{Open Digital Rights Language (ODRL)\footnote{\url{https://www.w3.org/TR/odrl-model/}}},  which is a W3C recommendation,  has gained a lot of traction in recent years in terms of intellectual property rights management (e.g.,~\cite{havur2019automatic, moreau2019license}).  Additionally, the ODRL model and vocabularies have been extended in order to model contracts \cite{guth2003experiences},  personal data processing consent \cite{esteves2021odrl}, and data protection regulatory requirements \cite{ de2019odrl}.  There has also been some work on automatically extracting rights and conditions from textual documents (e.g.,~\cite{cardellino2015information, DBLP:conf/esws/CabrioAV14}) or extracting important information from legal cases (e.g.,~\cite{wyner2010lexical, jurixNavas-Loro18}). \added{Although many of the proposed approaches are based on existing standards, there is a lot of overhead involved for systems that need to consider different types of policies that are encoded using different languages. General-purpose policy languages are particularly attractive in such scenarios as they lessen the administrative burden. However, considering the potential complexity of such a language, there is a need for policy profiles with well-defined semantics and complexity classes.}

\bigskip
\noindent
\added{\textbf{Policy Enforcement and Governance.}}
From a policy governance perspective, LegalRuleML researchers have proposed automated compliance approaches based on auditing (e.g.,~\cite{dimyadi2017evaluating, palmirani2018}) and business processes (e.g.,~\cite{palmirani2018legal, bartolini2019enhancing}). While \cite{governatori2016semantic} shows how LegalRuleML together with semantic technologies, is used for business process regulatory compliance checking based on a rule-based logic combining defeasible and deontic logic.  One of the advantages of description logic-based approaches, when it comes to consistency and compliance checking, is that they can leverage generic reasoners, such as Pellet\footnote{\url{https://github.com/stardog-union/pellet}} (e.g.,~\cite{francesconi2014description}).
Although there are presently no ODRL-specific reasoning engines, researchers have demonstrated how ODRL can be translated into rules that can be processed by Answer Set Programming (ASP)~\cite{baralbook} solvers such as Clingo~\cite{DBLP:journals/corr/GebserKKS14} (e.g.,~\cite{havur2019automatic, de2019odrl}).  Additionally,  there have been several custom applications that are designed to support ODRL enforcement or compliance checking, such as a license-based search engine~\cite{moreau2019license}; generalised contract schema and role-based access control enforcement~\cite{guth2003experiences}; and access request matching and authorisation \cite{esteves2021odrl}. \added{Despite existing efforts, challenges arise when it comes to ensuring that AI and processing algorithms adhere to the policies. This could potentially be achieved either before or during processing using trusted execution environments \cite{basile2023blockchain} or after execution by detecting data misuse via automated compliance checking using system logs \cite{kirrane2021specialk}. The combination of ex-ante and ex-post compliance checking is particularly appealing for supporting risk-based conformance checking such as that envisaged in the proposed EU AI Act. Nevertheless, the practicality, performance and scalability or these proposals remain to be determined.
%In both cases we assume an honours system whereby data processors are well intentioned and want to demonstrate compliance.
In order to further support self-determination, data owners and processors should be able to engage in on-demand negotiation over policies, assisted by technology that ensures a safe and fair space and helps assessing the compliance of negotiated terms with existing regulation. Negotiation between automated agents has been a topic of interest since the early 2000s but in the context of self-determination we must pay attention to the right balance between artificial representation and human involvement \cite{decision-support-negotiation,Baarslag2022}.}

\begin{figure}[t]
    \centering
    \includegraphics[width=\textwidth]{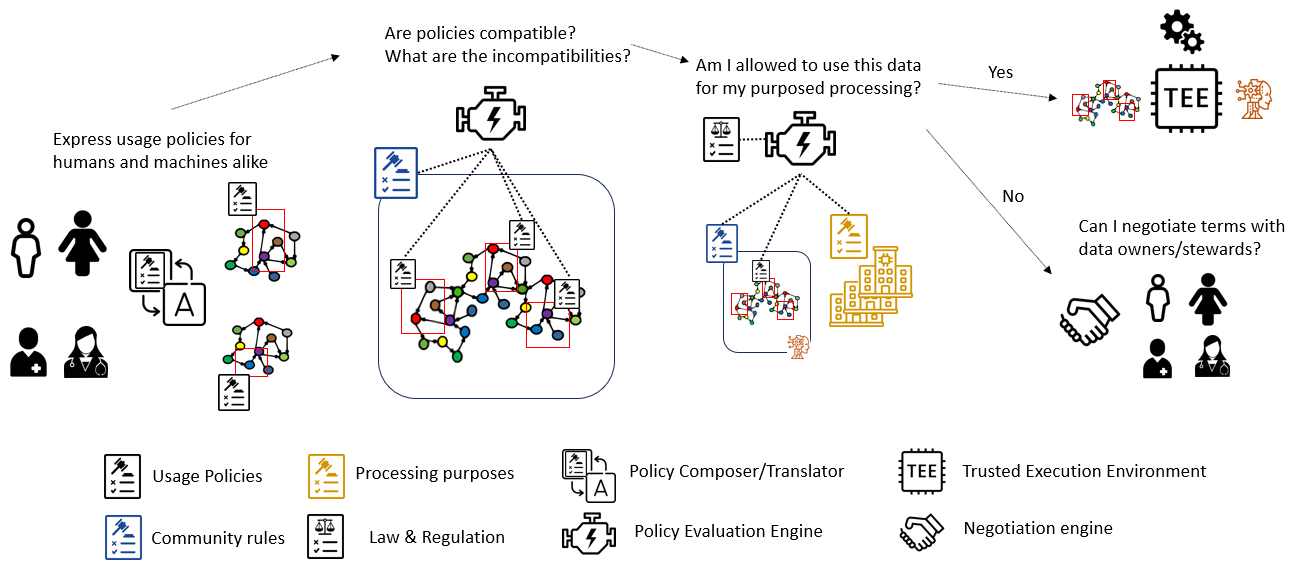}
    \caption{\textbf{Machine-readable norms and policies to support self-determination.} A Policy Composer/Translator assists individuals in writing data usage policies, communities in defining their rules, and organisations in declaring their data processing purposes in both human- and machine-readable formats. Policy Evaluation Engines assess the acceptability of perspectives in a community by evaluating policies and rules. Engines assess organizations' data usage compliance with regulations. If permitted, processing can occur in a Trusted Execution Environment ensuring compliance. If not allowed, a Negotiation engine may be utilised to seek agreement with data owners/stewards under relevant regulations.}
    \label{fig:MR-policies}
\end{figure}

\bigskip
\noindent
\added{\textbf{Grounding based on our Illustrative Scenario.}}
\added{Figure \ref{fig:MR-policies} illustrates how machine-readable policies and norms can be used to support self-determination.
Considering our illustrated scenario} individuals may want to establish policies to precisely define the subset of their PKGs to be shared with communities and what forwarding they allow. For example, \emph{share with the diabetes community my blood in sugar values measured by my connected device and the output of my AI healthcare assistant, or only share and forward anonymised aggregates to medical research institutions, or contact me for negotiation if the pharma company is interested in using my data for clinical studies.} Communities may do the same, \emph{e.g}, requiring \added{specific data to be shared} to join the community, but also \added{requiring agreements in order to ensure that participants will abide by} social and behavioural norms needed for self-regulation. Public and private organisations may need to adhere not only to privacy preferences and licenses but also to various \added{general regulations, e.g., the GDPR, the proposed AI Act in the EU or the Health Insurance Portability and Accountability Act (HIPAA)\footnote{\url{https://www.hhs.gov/hipaa/index.html}} in the US, as well as domain-specific regulations (e.g., advanced therapy medicinal products\footnote{\url{https://eur-lex.europa.eu/legal-content/EN/TXT/?uri=celex\%3A32007R1394}} and rare diseases\footnote{\url{https://eur-lex.europa.eu/legal-content/EN/TXT/?uri=celex\%3A32009H0703\%2802\%29}}).}

\subsection{Decentralised Infrastructure}
Over the last 15-20 years, a number of communities have come to accept that centralised computing systems, despite many benefits, can lead to issues such as the over-centralisation of power, the risk of single points of failure, potential abuse of personal data and creation of data silos which can inhibit innovation. A boon from this realisation is that we now have a number of technologies, standards, and approaches to decentralisation which offer benefits in terms of scalability, diversity, and privacy, as well as individually-centred flexibility and control, and is an appealing basis for maintaining and increasing trust, accountability, and autonomy with KG-based AI.

\bigskip
\noindent
\added{\textbf{Personal Knowledge Graphs.}} 
\added{The concept of a Personal Knowledge Graph (PKG), is that an individual can keep their personal or private data in a space belonging to them, rather than with siloed centralised service providers with limited access and control \cite{balog2019personal}. A Solid pod\footnote{\url{https://solidproject.org/}}} is an example of a PKG platform, and the key to the vision of Solid is that there should be standard interfaces and authorisation models to grant or deny access to the contents of a PKG at a granular level. This is argued in particular\footnote{\url{https://ruben.verborgh.org/blog/2017/12/20/paradigm-shifts-for-the-decentralized-web/}} to enable a highly decentralised architecture for Web applications. Rather than a provider aggregating data from all users into a single location controlled by the provider and application code accessing such data there, instead, an individual permits (or does not permit) Web applications of their choice to access whatever subsets of their data they decide from their PKG. As well as autonomy, this enables greater accountability since access to the PKG can be filtered via personal machine-readable policies at source, and activities can be tracked directly (e.g., \cite{domingue2023trusting}).
\added{Although PKGs offer great potential, they also come with challenges in terms of performance and scalability as applications will need to interact with multiple distributed data sources as opposed to a single backend server. These challenges, however, may also simultaneously be opportunities for scalability trade-offs, querying over multiple low-powered data sources rather than a high-powered central one.}

\bigskip
\noindent
\added{\textbf{Distributed Ledger Technology.}}
Distributed Ledger Technology (DLT) \cite{sunyaev2020distributed} promotes trust and empowerment through the replication of data across contributing nodes, which are geographically distributed across many sites, and the use of consensus algorithms which enable collective fair decision-making with no central control. Blockchains are a type of distributed ledger where an ever-growing list of records or blocks are tied together with cryptographic hashes, often, although not necessarily, associated with a securely exchangeable token system, or `cryptocurrency'. This technology rose to prominence following the release of Bitcoin \cite{nakamoto2008bitcoin} in 2008 - a blockchain-based currency that has now been adopted by El Salvador as their legal tender. Ethereum \cite{wood2014ethereum}, a blockchain platform released in 2015, contains the notion of a ‘Smart Contract’~\cite{buterin2014next} (originally coined in the 1990s by Nick Szabo~\cite{szabo1997formalizing}), which is a collection of code that executes in a fully decentralised way. %Specifically, being DLT data, the code itself is replicated across nodes, and every run of a smart contract is executed on every node, meaning that outputs can be cross-checked for reliability and good node behaviour. 
Smart Contracts have been used to implement a range of decentralised applications, including Decentralised Autonomous Organisations (DAOs) \cite{liu2021dao}, which are organisations where decisions are made through blockchain consensus mechanisms. The best-known example of a DAO was `The DAO' which at one point was worth more than \$70M; they have been applied to a number of different activities, including scholarly publishing \cite{hoffman2019scholarly}. \added{Despite the fact that immutability and transparency guarantees offered by DLT are very attractive, when dealing with personal data both the ledgers and the smart contracts themselves will need to be protected against unauthorised access and usage, and designed such that personal data itself is neither stored in, or derivable from, immutable DLT records. Smart contracts may also introduce scalability issues: the default Ethereum model involves every contributing node executing every run of a smart contract, and thus has inherent scale limitations. Relaxing this model may, however, affect trust.}

\bigskip
\noindent
\added{\textbf{Self Sovereign Identity.}}
In the Web space, Self-Sovereign Identity (SSI) is being developed through a combination of Decentralised Identifiers (DIDs) \cite{sporny_decentralized_2022} and Verifiable Credentials (VCs) \cite{sporny_verifiable_2022}, W3C standards for identity and verifiable attestation claims, respectively. DLT is one of the ways in which DIDs can be grounded, although, by design, the DID standard is open in terms of method. A DID is a URL (\texttt{did:<method>:<...>}) which can be resolved in a method-specific manner (e.g., HTTP(S) dereferencing, reading from a smart contract, etc.) to obtain a DID document, a Linked Data set containing information about digital identity in a standard form - for example, how to verify it (e.g., a public key), methods for communicating with the entity controlling it, and so on. DIDs enable SSI; the creation and use of DIDs are open and decentralised, and by using different DIDs with different audiences, individuals can minimise how easily their information can be tracked or correlated across services and can contextually and selectively disclose personal information as desired. This grants individuals significantly greater autonomy than current practices. There is a potential trade-off with trust and accountability of an individual when it comes to information that others need to rely on, which is that effective anonymity of a unique DID can be used to misrepresent oneself (e.g., fake a qualification or entitlement) or pretend to be someone else. VCs are a proposed solution to this. The VC data model is for sharing data alongside information that a recipient can use to verify its integrity or origin, such as a digital signature or DLT record. If a DID is presented to a service that is restricted to legal adults, for example, the DID owner may also present a VC issued by a government body confirming their adulthood; methods for selective disclosure supported by both DID and VC standards allow this to be done verifiably without requiring disclosure of real-world identity. \added{These technologies are relatively new in comparison with standard digital identity models, and, while intended and designed to address issues in those models, they may also introduce new difficulties or enable different vulnerabilities to, e.g., identity fraud, than current standards.}

\bigskip
\noindent
\added{\textbf{Federated Learning.}} 
In the context of data-driven AI and decentralised infrastructure, there are also techniques for decentralised machine learning. Federated Learning (FL) \cite{yang2019federated} is the idea that, rather than aggregating training data in one location controlled by a model developer (thereby compromising subject privacy), data holders can run learning algorithms to generate model weights for their own data locally and privately, and then send only the weights to the developer to be incorporated into the larger model. An example might be a smartphone text prediction personalisation algorithm, where a user’s own writing is used to generate predictive weights on device, and periodically selections of these can be aggregated to improve general text prediction models. Refinements of FL approaches include sending not the actual learned model weights, but a set of weights with statistically similar properties \cite{wei2020federated}, to further reduce the risk of privacy breaches without affecting model performance. A related approach takes this concept even further, with the idea of embeddings in a larger model, e.g.,     `Textual Inversion' \cite{gal2022image} to personalise large generative image diffusion models. The intuition here is that if someone wants certain personalised specific types of output from a generative AI, then, if a model is sufficiently large, there is a good chance that the desired concept already exists within it. \added{More recently, the idea of federating for preserving privacy has been applied specifically to deep learning, in particular in the context of Internet of Things. \cite{Yin2020} proposes an architecture with a control layer including a distributed ledger, while \cite{Xu2022} propose advanced cryptographic mechanisms to reduce the risk of privacy leaks, following more general approaches that apply either differential privacy, homomorphic encryption or secure multi-party computation. Federation also has the positive side effect of potentially speeding up model training when the privacy constraints allow for a helpful distribution of the process \cite{Ben-Nun2020}. However, when opening the process to multiple parties, there are a number of attack vectors that do not exist in a centralised approach for which we need protection, and pay a communication and computation overhead \cite{Hu2022}}.  %For example, while it might not have been trained on any images of a particular dog, it will have been trained on enough dog images that there will be some input which can lead it to generate images which look extremely like that dog. Given a small number (5-6) of images of the dog in question, it is possible to learn what that input embedding is, and, combining that with other inputs, generate novel images which appear to show the same dog. The same technique can be applied to generative models in general. This approach does not modify the model itself, or require that the example data or resultant embeddings are shared, and it is computationally cheaper than training the model further. It therefore offers the potential, in certain circumstances, for fully private decentralised personalisation of AI models.

% \begin{figure}[t!]
%     \centering
%     \includegraphics[width=0.7\textwidth]{images/decentralised-infrastructure.png}
%     \caption{Decentralised Infrastructure supporting self-determination, shown from the perspective of one individual with a PHKG\footnote{The full picture would have knowledge exchange between multiple parties; to avoid an unreadable cluttered figure, this is left implied by the background network.}. Selected data from the PHKG can be shared with different audiences according to individual wishes, either directly with a healthcare provider, with Web applications for health, or indirectly with peer or research communities. Identity is via DIDs (anonymous in the latter cases), with VCs used for trustable selective disclosure. KG-based AI models can be trained and personalised in federated and private ways on knowledge from diverse sources}
%     \label{fig:decentralised-infrastructure}
% \end{figure}

%\begin{wrapfigure}{r}{0.68\textwidth}
%  \vspace{-20pt}
% \begin{center}
%\includegraphics[width=0.66\textwidth]{images/decentralised-infrastructure.png}
\begin{figure}[t]
    \centering
    \includegraphics[width=0.75\linewidth]{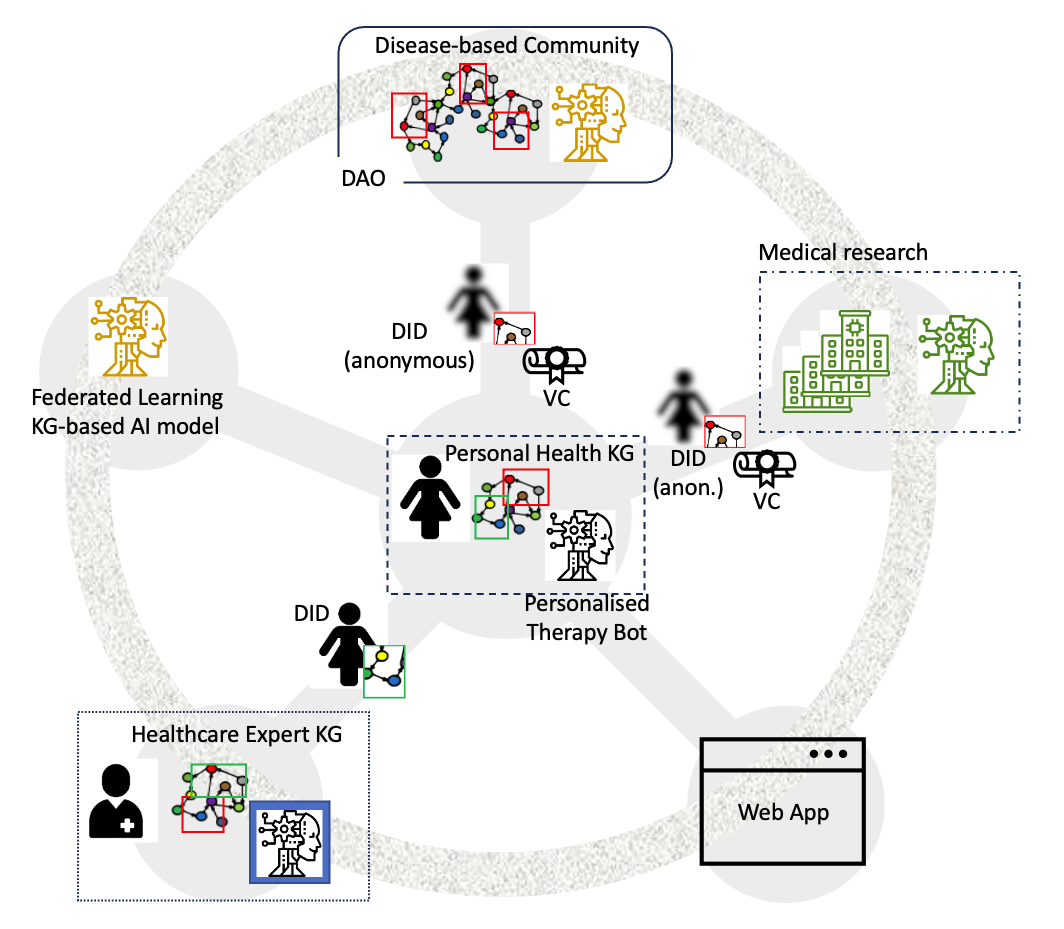}
%\end{center}
%\vspace{-20pt}
    \caption{Decentralised Infrastructure supporting self-determination, shown from the perspective of one individual with a PKG\protect\footnotemark. According to individual wishes, portions of the PKG can be shared either directly with a healthcare provider, with web applications for health, or indirectly with peer or research communities. Identity is via DIDs (anonymous in the latter cases), with VCs used for trustable selective disclosure. KG-based AI models can be trained and personalised in federated and private ways on knowledge from diverse sources.}
    \label{fig:decentralised-infrastructure}
\end{figure}
%\vspace{-30pt}
%\end{wrapfigure}
\footnotetext{\added{The full picture would have knowledge exchange between multiple parties; to avoid an unreadable cluttered figure, this is left implied by the background network.}}
\bigskip
\noindent
\added{\textbf{Grounding based on our Illustrative Scenario.}}
\added{A decentralised infrastructure supporting self-determination for our illustrative scenario is depicted in \autoref{fig:decentralised-infrastructure}}. Health data is highly sensitive and private, and individuals may want or need to interact with multiple services where it is relevant, including KG-based AI systems. It thus makes sense to create a personal health knowledge graph (PKG) to be a comprehensive and interconnected representation of an individual’s health information, including their medical history, lifestyle choices, genetic data, and real-time health monitoring data from IoT devices. Data from various sources, such as wearable devices, mobile applications, electronic health records, and even genomic sequencing, can be linked together to form a holistic view of an individual’s health in such a personal health knowledge graph. An early example of a PKG was in \cite{third2016integrating}, where medical, lifestyle, and IoT health monitoring data in a PKG was integrated into a (patient-focused) decision support system built around a public medically-curated KG representing cardiovascular risk factors, giving individuals the autonomy to gain deeper insights into their own health patterns and risks, identify correlations, and make more informed decisions. 
%For example, if an individual’s heart rate data from their wearable device, coupled with their medical history and lifestyle choices, indicates a higher risk of cardiovascular disease, they can receive evidence-based lifestyle or medical recommendations. However, the knowledge that fuels such applications may come from various decentralised sources, which may have different levels of trust metrics associated. Therefore, when such knowledge is utilised in an AI-based health diagnosis, we need to have mechanisms that can verify the authenticity of the data, and have mechanisms to streamline clinical decision support systems that use it. 
More recently, BlockIoT~\cite{shukla2021blockiot,shukla2021blockiotretel} aims to integrate health data seamlessly in a decentralised PKG using blockchain and KG technologies, addressing this trust aspect and using PKG-driven smart contracts to trigger the personalised recommendations for lifestyle modifications, medication adjustments, or even timely interventions by the healthcare providers. Furthermore, the PKG can serve as a powerful tool for healthcare beyond the individual. Communities of patients, providers, researchers, etc., or combinations thereof, can share knowledge about various aspects of, e.g., particular conditions, whether that is clinical evidence and best practice, peer advice and support on living with a condition, or data on novel or rare symptoms and side effects, with this knowledge used for support, care, or medical research across populations. De-identified and aggregated data from multiple individual KGs can be collected in community KGs, with trust securely established using DIDs and VCs, and accessed by community, practitioner, researcher, and service provider stakeholders, allowing for decentralised large-scale analysis and identification of broader health trends from multiple perspectives and intersecting factors. This can lead to advancements in disease prevention, treatment protocols, and the development of personalised medicine in a collaborative manner~\cite{shukla2022collaboratively}.  
KG-based AI systems can be both trained and used across this ecosystem, with FL being applied to train larger models (e.g., the organisation models in \autoref{fig:scenario}) and personalised embeddings used by individuals to get the best experience from their therapy bots and healthcare assistants while maintaining privacy and autonomy. 

\subsection{Decentralised KG Management}

As the amount of data and knowledge grows exponentially, managing and harnessing this vast information becomes increasingly complex. Traditional centralised approaches to KG management face challenges in terms of scalability, privacy, and control over data, and to address these issues, decentralised KG management emerges as a promising solution. This section explores the key aspects and open challenges in decentralised KG management to enable trust, accountability, and self-determination for individuals in a rapidly evolving AI ecosystem. 
%Setting up decentralised KGs requires transforming multi-modal data about a person in a PKG into factual statements representing relational properties and meaning. Moreover, data and metadata should be fit for use, meet specific quality standards, and validate integrity constraints; engines should be capable of distributing KGs to respect regulations for data access and sharing. Traversing decentralised KGs demands infrastructures capable of efficiently executing queries against the available KGs, while respecting norms and policies. Lastly, KG management tasks (i.e., data curation, integration, validation, distribution, and exploration) should be traceable and validatable giving transparent management pipelines for self-determined KG-based AI~\cite{geisler_knowledge-driven_2022}. 

\bigskip
\noindent
\added{\textbf{Decentralised KG Access and Management.}} 
Efficient query processing infrastructures are fundamental for traversing decentralised KGs. There has been notable efforts such as Fedbench~\cite{schmidt2011fedbench} in the past. However, these infrastructures should be capable of executing queries against the available KGs while respecting privacy and adhering to norms and policies. With the increasing emphasis on privacy protection with regulations such as GDPR, it is crucial to develop mechanisms that allow users to access and extract knowledge from KGs without compromising sensitive information or violating privacy regulations.
Several research directions are worth considering to address the open challenges in decentralised KG management. Firstly, developing the formalisms to describe KG management semantically can provide a common ground for understanding and interoperability across different decentralised KG systems. Such formalisms can enable standardised representations of KGs in the form of ontologies and facilitate seamless integration and collaboration among diverse knowledge sources. Architectures supporting new protocols and standards specific to decentralised KGs are essential for establishing interoperability and seamless communication between knowledge sources and systems. By defining and adopting common protocols and standards, decentralised KGs can collaborate more effectively, share insights, and facilitate cross-domain knowledge discovery.

\added{Note that if we add LLMs to the picture, their current training and execution processes are currently centralised. Decentralised KG management is useful to provide transparency in data used for their training. For approaches involving the interaction between LLM and KGs, the transparency of the LLM itself still depends on the owner.} 

\bigskip
\noindent
\added{\textbf{Provenance and Explanations.}} 
Furthermore, explainable methods for data integration and curation, as well as KG validation and distribution, such as the \emph{Explanation Ontology} for user-centric AI, are necessary to ensure the reliability and accuracy of decentralised KGs~\cite{chari2020explanation}. By providing transparent and interpretable approaches, users can have better insights into knowledge integration and validation, enhancing trust and accountability \added{of the knowledge contained in the KG and the insights derived. This is especially critical because, in decentralised KGs, data may come from various sources and be represented in different ways. The standardised framework provided in the Explanation Ontology for representing domain-specific explanations of KG entities and relationships helps users and applications understand the meaning and context of the data in the KG.}
Provenance and traceability \added{also play a vital role in decentralised KG management. Establishing} mechanisms to track and validate the origin, history, and lineage of knowledge within KGs is crucial for accountability and the ability to trace back the sources and transformations that contribute to the resulting knowledge. The W3C Provenance Data Management standards~\cite{missier2013w3c} provides the basis for encoding provenance attributes in KGs, and subsequent nanopublications specification~\cite{groth2010anatomy} has gained a lot of traction in the biomedical domains. \added{While these solutions exist, there needs to be a cohesive framework that ties together explanation provenance data management in a decentralized KG context, ensures that users can trace the origins, transformations, and sources of the data, which is crucial for trust, accountability, and data quality assurance. 
The W3C provenance data management suite of recommendations provides normative interoperable guidance on recording information about data sources, contributors, and how data is collected or transformed, making integrating heterogeneous data into a coherent KG easier. When data quality issues arise, users can trace back to the source of the problem and take corrective actions, ensuring the KG remains accurate and reliable. The W3C recommendations for decentralized provenance management provide a mechanism for attributing data to its sources or contributors. This attribution is essential for accountability, especially when multiple parties contribute to a~KG.}

\bigskip
\noindent
\added{\textbf{Blockchain Technologies and Tokenomics.}}
In recent years, the integration of blockchain technologies and tokenomics has gained attention in the context of decentralised KG management. Projects such as OriginTrail\footnote{\url{https://origintrail.io}} have contributed to the development of ownable DKGs, which leverage blockchain's inherent properties to enhance trust, provenance, and accountability. By utilising blockchain, KG management systems can ensure the integrity and traceability of data and metadata across various nodes in the network.
The OriginTrail protocol aims to create a trustless environment where data providers, consumers, and verifiers can interact and validate the authenticity and reliability of data stored within the knowledge graph. Their protocol issues tokens as incentives for data contributors, validators, and curators within the KG ecosystem.
% This token-based economy incentivizes active participation and rewards users for their contributions to data curation, validation, and exploration processes, which fosters a community-driven and self-sustainable ecosystem that promotes trust, transparency, and accountability. 
The integration of blockchain technologies and tokenomics in decentralised KG management addresses several critical aspects. Firstly, blockchain's immutability and transparency enable the traceability and provenance of data and metadata, ensuring accountability throughout the KG management pipeline. Secondly, the decentralised nature of blockchain mitigates single points of failure and promotes the distribution of knowledge and decision-making power among participants. This decentralised approach aligns with the principles of self-determination, empowering individuals to have control over their data and knowledge.
By rewarding contributors, validators, and curators with tokens, these systems encourage continuous improvement, data quality assurance, and community engagement. Token-based economies can drive the development of sustainable KG management pipelines, enabling the growth and evolution of DKGs over time.  \added{However, the tokenomics have to be carefully designed and monitored to avoid the possibility contributors have a motivation (possibly extrinsic) to misbehave. There is also the risk that a sudden churn in blockchain participants impacts performance and availability. There is also the question of the performance of the consensus algorithm of the Blockchain itself.}

\begin{figure}[t]
    \centering
    \includegraphics[width=0.9\linewidth]{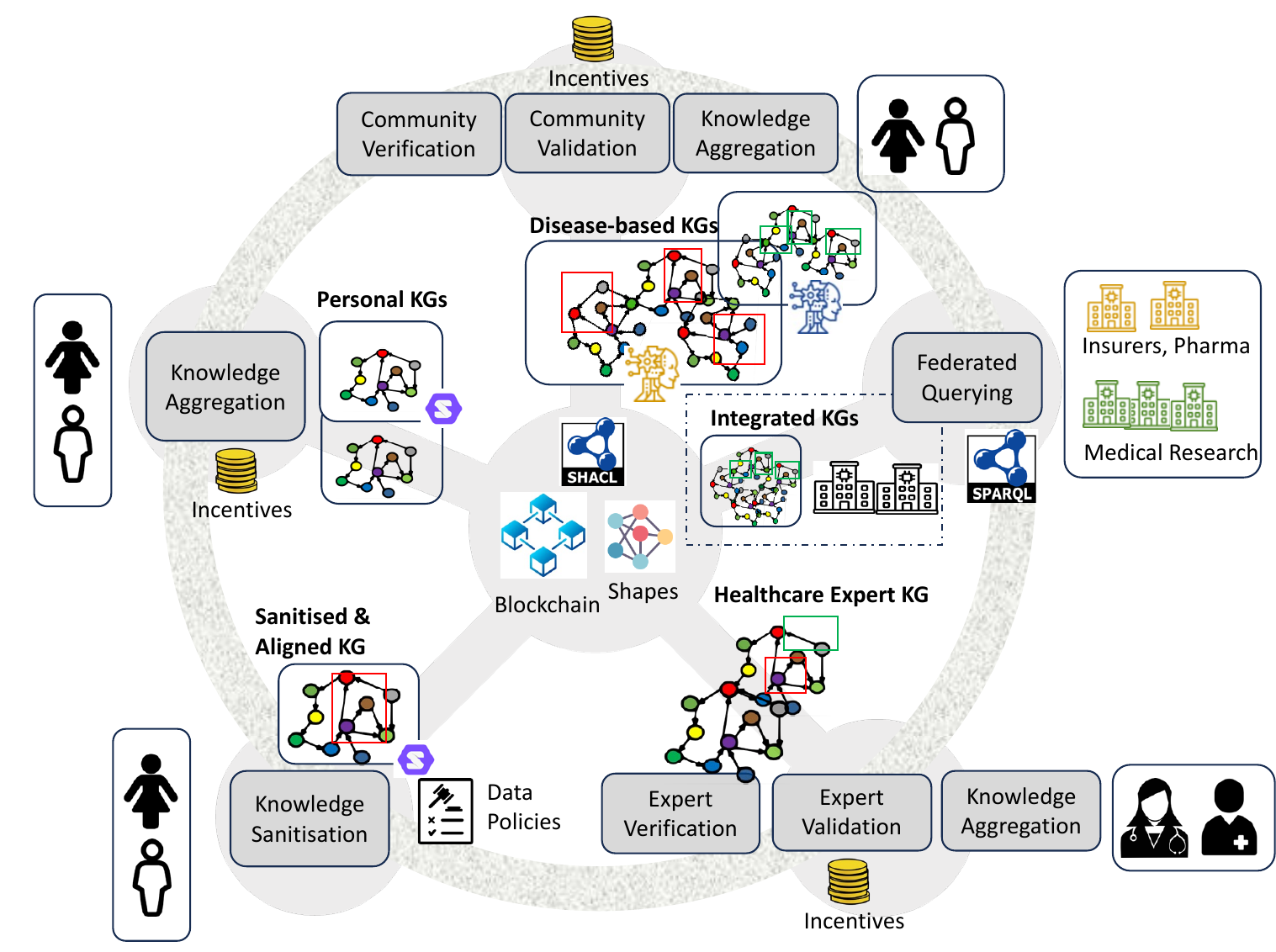}
    \caption{\textbf{Decentralised KG Management Process in Healthcare.} Emphasising user empowerment, privacy, and seamless collaboration, users maintain control over their personal health data through personal data stores like Solid, and community and healthcare experts enhance different facets of the KGs in the ecosystem. decentralised KG management involves anonymisation, filtering based on data policies (including GDPR and HIPAA), and alignment with community-defined ontologies. Incentives, driven by blockchain technology, encourage user participation in aggregating KGs and incentivize healthcare professionals for verification, validation, and aggregation activities. SHACL shapes ensure KG validation and federated querying mechanisms enable access to the KGs to stakeholders, e.g., insurers, pharma, and medical research organisations. Integrated KGs are iteratively generated; they comprise a federation of KGs that may be autonomous, distributed, and heterogeneous. A federation query engine enables the traversal of these integrated and connected KGs to provide useful insights to the stakeholders involved.}
    \label{fig:dkg-management}
\end{figure}

\bigskip
\noindent
\added{\textbf{Grounding based on our Illustrative Scenario.}}
\added{An approach to decentralised knowledge graph management in the context of healthcare, where users retain control over their personal information while benefiting from enhanced privacy measures and seamless collaboration in a community, is illustrated in \Cref{fig:dkg-management}}. At the heart of this framework lies the concept of PKGs, such as Solid, which empower individuals to securely store and manage their personal health data. Central to the architecture are specific components aimed at safeguarding user privacy and ensuring data transparency. The process begins with knowledge sanitisation, which anonymises sensitive information and filters the data according to the user's preferences and data policies. These policies encompass not only globally recognised regulations like GDPR and HIPAA but also individual data policies, enabling users to set granular restrictions on how their data is used, such as opting out of genetic data usage for medical research.
To ensure interoperability and standardisation, the creation of knowledge graphs leverages community-defined ontologies and vocabularies. These shared frameworks facilitate seamless integration and alignment of personal knowledge graphs within the broader ecosystem, promoting data exchange and collaboration.
Users are incentivised to aggregate their knowledge graphs, contributing to the construction of community-based knowledge graphs focused on specific diseases. Through community-based verification, validation, and knowledge aggregation processes, these disease-based knowledge graphs are created, providing valuable insights and fostering collaborative efforts among healthcare professionals, researchers, and the wider community. Blockchain-based incentives drive user participation, rewarding both community users and healthcare experts for their verification, validation, and aggregation activities. The utilisation of an immutable ledger and verifiable credentials ensures the integrity and trustworthiness of the verification process. The validation process, powered by RDF SHACL and Shape descriptions, further enhances data quality and consistency, instilling confidence in the aggregated knowledge.
The integrated knowledge graphs, encompassing personal, community-based, and healthcare expert knowledge, can be queried using federated querying mechanisms powered by SPARQL. This allows various institutions, including insurers, pharmaceutical companies, and medical research organisations, to access and leverage the rich insights stored within the knowledge graphs, enabling evidence-based decision-making and advancing medical research and healthcare practices.
By combining decentralised knowledge graph management, user-centric privacy controls, and collaborative data sharing, this innovative framework represents a significant step forward in transforming decentralised KG management, fostering a secure, privacy-enhanced environment that empowers users, facilitates collaboration, and drives advancements in domains such as medical knowledge and patient care.

%Although the term Decentralised KG Management was mentioned as early as 1972 \cite{schneider_course_1973}, most researchers associate the term with the announcement by Google in 2012 of their own Knowledge Graph built on top of public sources independently controlled, such as DBpedia and Freebase. In turn these two technologies are founded on RDF, a cornerstone of the Semantic Web and Linked Open Data research areas. The commonality in all these activities is the presence of a machine-readable syntax for data and data schemas based on a web standard.  Cognitive theories conceptualise human decisions as the cooperation of two main types of capacities, the ones providing intuitive and unconscious choices and the capabilities to manage logical and rational thinking. AI statistical and symbolic methods emulate these human capacities and, independently, have contributed to many real-world applications. However, unifying these approaches in transparent frameworks is still challenging, and research on neuro-symbolic systems is an emerging trend to empower AI models with both types of abilities naturally included in human intelligence \cite{breit_combining_2023,rivas_neuro-symbolic_2022}. 

\subsection{Explainable neuro-symbolic AI}

Neuro-symbolic systems go beyond generating explanations solely based on the trained model or the individual results derived from applying the model to specific data. They can produce symbolic explanations capturing the essence of an AI model itself. These explanations can be classified as either \emph{instance-level} explanations generated for each specific result of the model, or \emph{model-level} explanations of the structure of a learned model. Previous work on the role of KGs in AI has focused on explainability. \cite{lecue_role_2020} frames explainability as a dimension of \emph{trustable} AI and presents  challenges, existing approaches, limitations and opportunities for KGs to bring explainable AI to the right level of semantics and interpretability. \cite{tiddi_knowledge_2022} and \cite{rajabi_knowledge-graph-based_2022} conducted independent systematic reviews of existing explainable AI systems to characterise KGs' impact. These results put into perspective the role of KGs in providing symbolic reasoning and learning capabilities with the potential to be precise-- as shown by Akrami et al. \cite{DBLP:conf/sigmod/AkramiSZHL20}-- in addition to being explainable. 

\bigskip
\noindent
\added{\textbf{Reasoning and AI.}}
\added{Despite the unquestionable reasoning features of symbolic systems and the studies reporting limitations of LLMs in human-like tasks (e.g., explanations, memories, and reasoning over factual statements) \cite{DBLP:conf/nesy/HammondL23}, and there is an ongoing debate about LLM's reasoning their causal inference capabilities \cite{DBLP:journals/corr/abs-2305-00050}. 
Although LLMs excel at certain reasoning tasks, they do poorly in others, raising the question if they genuinely engage in causal reasoning or merely function as unreliable mimics, generating memorized responses (e.g., \cite{huang-chang-2023-towards}). Methods to reason can be roughly divided into methods using only the LLM itself (e.g. with prompt-engineering), and methods combining the LLM with an external reasoner and/or external source of knowledge (e.g. a Knowledge Graph) \cite{qiao-etal-2023-reasoning}. Our vision posits that external help will always be needed, especially for concrete use cases.}    
\added{There are discussions about the need for knowledge graphs in the era of LLMs. Sun et al. \cite{DBLP:journals/corr/abs-2308-10168} and Dong \cite{DBLP:journals/corr/abs-2308-14217} report on an empirical assessment of ChatGPT~\cite{schulman2022chatgpt} with respect to DBpedia, illustrating the need of symbolic systems that \emph{over-fit for the truth} whenever factual statements are collected from KGs. In addition, symbolic approaches can support sanity checking and be easily auditable and traceable. These features position the combination of both approaches in neuro-symbolic AI as a feasible option to provide KG-based AI. The neuro-symbolic AI delivers the basis to integrate the discrete methods implemented by symbolic AI with high-dimensional vector spaces managed by LLMs. They must decide when and how to combine both systems, e.g., following a principled integration (combining neural and symbolic while maintaining a clear separation between their roles and representations) or integrated (e.g., a symbolic reasoner integrated into the tuning process of an LLM).}
%Recently, van Bekkum et al. \cite{DBLP:journals/apin/BekkumBHMT21} propose 17 fundamental design patterns to model neuro-symbolic systems. These patterns encompass many scenarios where the symbiotic relationship between symbolic reasoning and ML models becomes apparent. Since these combinations may enable symbolic reasoning and enhance contextual knowledge, neuro-symbolic systems may empower explainability and, as a result, also improve transparency by showing how a system works based on the symbolic explanations deduced by the hybrid system. }

\bigskip
\noindent
\added{\textbf{Trust and AI.}}
Trust in AI systems stems from various factors, including transparency, reproducibility, predictability, and explainability. Neuro-symbolic systems play a vital role in enhancing trustworthiness by enabling communication between modules and facilitating tracing. Modularity enables the specification, verification, and validation of each component and its interactions. As a result, a system's behaviour can be traced and validated.
Specifically, within the domain of KG-based AI for self-determination, the seamless integration of KGs and symbolic semantic reasoning offers a comprehensive and unified perspective on curated knowledge. This integration holds immense value in addressing critical tasks such as validating, refuting, and explaining incorrect, biased, or misleading information that may potentially be generated by LLMs. By combining symbolic reasoning over KGs with LLMs, the propagation of misinformation can be mitigated while simultaneously enhancing the transparency and trustworthiness of AI-generated outputs. Consequently, KG-based AI systems can effectively emulate human behaviour by subjecting mistakes arising from false or incomplete information to a process of validation and enrichment using curated and potentially peer-reviewed sources of knowledge \cite{DBLP:journals/apin/BekkumBHMT21}.

\bigskip
\noindent
\added{\textbf{Quality and AI.}}
A notable application of KGs in neuro-symbolic AI is as a source of informative prior knowledge to increase the quality of machine learning models. An example is the work by Rivas et al. \cite{DBLP:journals/semweb/Rivas23}, where a deductive database, expressed in Datalog, establishes an axiomatic system of the pharmacokinetic behaviour of a treatment's drugs and enables the deduction of new drug-drug interactions in cancer treatments. This prior knowledge plays a crucial role in elucidating the characteristics of a therapy and justifying its efficacy by considering all the interactions and the dynamic movement of drugs within the body. It encompasses factors such as the absorption, bioavailability, metabolism, and excretion of drugs over time.
A KG embedding model improves its prediction of the effectiveness of a treatment, based on the prior knowledge which encodes statements about a treatment's characteristics; these statements are inferred by a deductive system which comprises the symbolic component of the hybrid approach. An approach for explaining link prediction (e.g., \cite{DBLP:conf/sigmod/0002FMT22}) allows the justification of why this added prior knowledge affects the model's decisions, potentially improving trust on the model's results. 

\begin{figure}[t]
    \centering
        \includegraphics[width=1.0\linewidth]{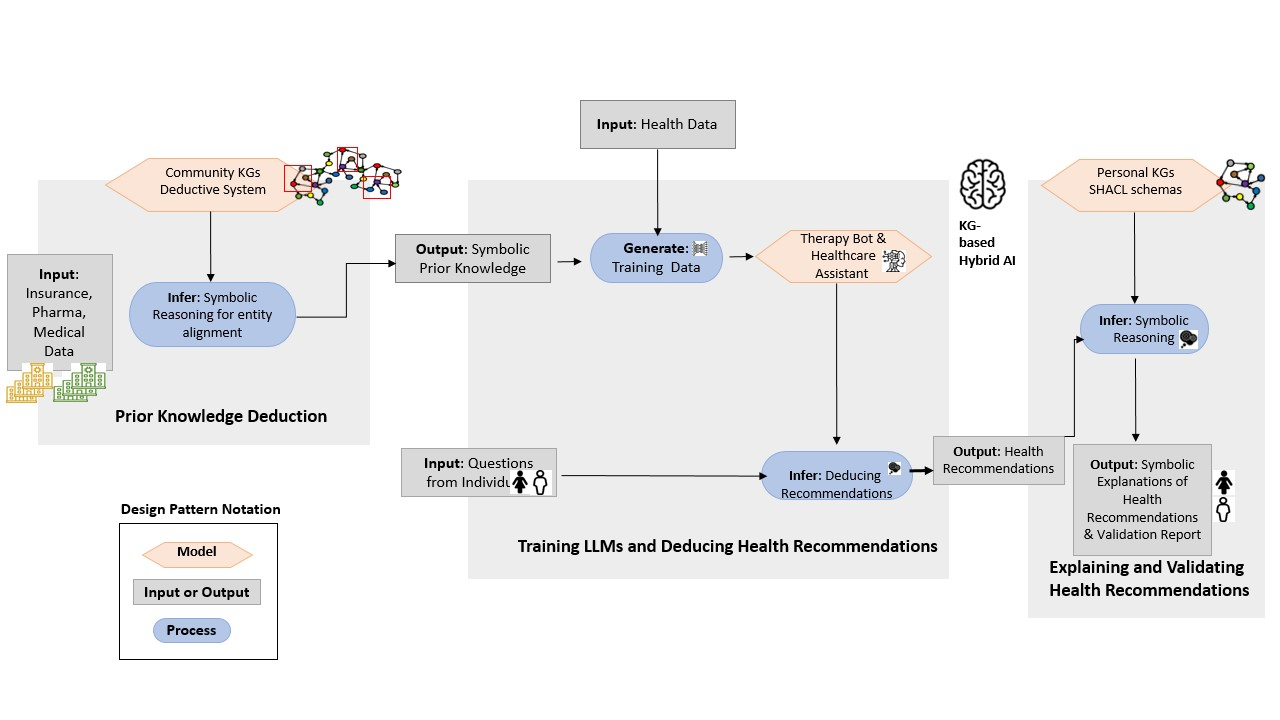}
        \vspace*{-5mm}
    \caption{\textbf{Design Patterns for Hybrid AI.}  
    Extension of patterns by van Bekkum et al. \cite{DBLP:journals/apin/BekkumBHMT21} for running example in \autoref{fig:scenario}. The patterns represent an explainable system with prior knowledge created by the alignments of data from health-related data sources (e.g., insurance, pharma, and medical data).}
    \label{fig:patterns}
\end{figure}
\iffalse
\begin{figure}[h!]
    \centering
    \subfloat[An Explainable AI Hybrid System Through Rational Reconstruction]{
        \includegraphics[width=1.0\linewidth]{images/pattern1.png}
        \vspace*{-5mm}
        \label{fig:pattern1}
    } ~\\\vspace*{.75em}
    \subfloat[An Explainable AI Hybrid System with Prior Knowledge]{
        \includegraphics[width=1.0\linewidth]{images/pattern2.png}
        \vspace*{-5mm}
        \label{fig:pattern2}
    }
    \caption{\textbf{Design Patterns for Hybrid AI.}  
    Extension of patterns proposed by van Bekkum et al. \cite{DBLP:journals/apin/BekkumBHMT21} to meet requirements of running example in \autoref{fig:scenario}. \autoref{fig:pattern1} depicts a pattern about an explainable learning system that resorts to PKGs, SHACL validation, and symbolic reasoning for explaining health recommendations generated by a therapy bot based on an individual's question. \autoref{fig:pattern1} illustrates a pattern for empowering an explainable system with prior knowledge created by the alignments of data from health-related data sources (e.g., insurance, pharma, and medical data).}
    \label{fig:patterns}
\end{figure}
\fi

\bigskip
\noindent
\added{\textbf{Grounding based on our Illustrative Scenario.}}
Grounding on the example presented in \autoref{fig:scenario}, when individuals and professionals engage in communities with bots and assistants powered by AI models, it is critical to ensure the transparency of their decision-making process. However, despite the increasing focus on LLMs in healthcare and their continual improvement in terms of precision and accuracy \cite{DBLP:journals/corr/abs-2305-09617}, their outcomes can still be susceptible to hidden biases and a lack of traceability \cite{Li2023}.
To tackle these challenges, the utilisation of a neuro-symbolic system can enhance LLMs by incorporating reasoning capabilities. This system operates as a deductive system on a user's Knowledge Graph (KG). 
These hybrid AI systems can be effectively modelled using patterns proposed by \cite{DBLP:journals/apin/BekkumBHMT21}. 
\autoref{fig:patterns} depicts a pattern describing a hybrid AI system that enhances the explainability of the LLMs described in our running example. 
At the community level, symbolic reasoning applied to the ontology of shared PKGs can generate prior knowledge, enabling precise and concrete questioning of an LLM and providing additional contextual information. Moreover, a symbolic system facilitates the linking of shared PKGs with corresponding entities in KGs related to insurance, pharmaceuticals, and medical research. By incorporating this prior knowledge, the LLM's answers are improved and validated with the assistance of the symbolic system. The systems operating at the community level and involving heterogeneous sources can be described using the \emph{explainable system with prior knowledge} pattern; data alignments comprising prior knowledge enhance contextual knowledge provided to the therapy bot, facilitating thoughtful health recommendations. 

\section{\added{Proposed KG-based AI for Self-determination Research Agenda}}
\label{sec:roadmap}

\added{In this section, we derive a set of requirements concerning KG-based AI for self-determination and map them to the concrete research goals introduced at the start of this vision paper.} 

\subsection{Trust, Accountability, and Autonomy Foundational Goals}

\added{In the following, we highlight five open research challenges and opportunities in each of our proposed foundational topics (machine-readable norms and policies; decentralised infrastructure; decentralised KG management; and explainable neuro-symbolic AI). Considering the complex nature of each of these requirements, an assessment of the maturity of existing technologies with respect to the various requirements is beyond the scope of a vision paper.} %\added{Additionally, an indicative overview of the maturity of the state of the art, based on the existence of conceptual work (models), practical instantiations (technologies), and broader engagement (standards), is provided.}

%\sabrina{My proposal to indicate the maturity is to rename short-term, medium-term and long-term to models, technologies, and standards such that we can give an indication of what is there already. For instance, we could potentially use a different number of *'s to indicate maturity. Clearly this is subjective, therefore if anyone has other idea's I'm all ears.}

\bigskip
\noindent
\added{\textbf{Machine-readable Norms and Policies.}}
\smallskip
\begin{description}
    \item[MRP1: Seamless policy translation.] \added{There is a need for humans to express policies in machine-readable format and for machines to express them in natural language or via appropriate visualisations. A major challenge involves checking that machine readable policies faithfully represent their human readable counterpart.}
    \item[MRP2: Multi-level policy evaluation.] \added{Several policy languages exist, however many of them do not have corresponding enforcement mechanisms. Given that usage constraints, community rules, and regulations operate at different, yet interconnected levels, there is a need to devise effective and efficient enforcement and/or compliance checking strategies.}
    \item[MRP3: Negotiation.] \added{Facilitate autonomy via fair and safe negotiation between individuals, communities, and organisations. Here there is a need to study the benefits and tradeoffs between merely assisting humans in taking decisions and developing automated approaches that alleviate individuals from constant affirmations (e.g., the cookie problem).} 
    \item[MRP4: Compliance verification.] \added{Provide support for both ex-ante and ex-post compliance checking mechanisms. Despite their potential, it remains to be seen which machine-readable agreements can actually be enforced by trusted execution environments. Additionally, in scenarios where it does not pay data processors to cheat, game theoretic approaches could be used to underpin honours based compliance checking.} 
    \item[MRP5: Data misuse detection.] \added{Instil trust and to ensure accountability in KG-based AI, by developing mechanisms that can detect if any party violated policies and norms. In this context, causal reasoning and explanations could potentially be used to both detect misuse and to better understand the root cause.}  
\end{description}

\smallskip
\noindent
\added{\textbf{Decentralised Infrastructure.}}
\smallskip
%
%As mentioned previously, numerous technology elements related to decentralised infrastructures align with our three pillars in terms of desires and requirements. However, these mechanisms are still in the early stages of development, resulting in ongoing research challenges.
\begin{description}
\item [DI1: Comprehensive recording.] A DLT can provide an immutable ledger but work remains on how best to connect KG-based AI activities, e.g., to a possible federated query engine. %[trust short term]
\item [DI2: Personalised tracing.] Providing individual and community owners of PKGs with personalised traces of how acquired data was processed and used, will involve dis-aggregating KG-processing and inferencing according to different user data and ensuring that privacy is not violated when individual results are returned. %An important first step is to enable organisations to publish anonymised summaries of activity allowing for high-level validation of `good' behaviour. %[trust and accountability Technologies]
\item [DI3: `Decency' check.] There is a need for easy-to-use services which allow users and communities to check if an organisation has behaved in a `decent' way when it processed acquired data. Research here will examine how `decency' can be defined and validated by comparing PKG declarations of use (e.g., policies) with generated traces of use. %[trust and accountability long term]
\item [DI4: Interoperability.] Develop mechanisms that facilitate comprehensive interoperable identification of human and machine participants in KG-based AI processes. For example, users and communities will wish to know, and be able to validate, claims that a data request comes from a particular organisation, unit and even individual KG processor. This will provide a foundation for accountability at all levels of granularity. %[trust and accountability short term]
\item [DI5: Self-sovereignty.] True self-sovereign KG-based AI needs to be: (i) based upon easy-to-use self-sovereign identities and data management; and (ii) capable of supporting the continuous monitoring of organisational behaviours in a transparent fashion.
\end{description}

\smallskip
\noindent
\added{\textbf{Decentralised KG Management.}}
\smallskip
%
%Proper Decentralised KG management is necessary not only to ensure transparency, explainability, and trustworthiness of AI processes but also to ensure the Standards sustainability of the knowledge that drives the AI. However, despite the existing advancements in various knowledge management infrastructures, we put forth several goals required to implement a cohesive Decentralised KG management that combines various complementary aspects to contribute to the three pillars of KG-based AI for self-determination depicted in \autoref{fig:pillars}, i.e., Trust, Accountability, and Autonomy, and these goals are defined based on the motivating use case illustrated in \autoref{fig:scenario}.
%
\begin{description}
\item [DKG1: Knowledge Sanitisation.]
Develop robust techniques for knowledge sanitisation that ensure user privacy by anonymising and filtering sensitive information based on data policies. These policies can be regulations such as GDPR and HIPAA, as well as individual-level data policies enforced at their personal data store, empowering users to specify their sharing preferences and control the aspects of data they disclose. %[Models]

%\item [DKG2: Alignment with Standardised and Community Ontologies.]
%Investigate approaches to align personal knowledge graphs with community-defined ontologies and vocabularies. The decentralised apps should leverage such ontologies. However, there may be some discrepancies in the ontologies used in the KG generation. In that case, we need seamless integration and interoperability mechanisms, enabling effective data exchange, collaboration, and knowledge sharing within the decentralised ecosystem. %[Models]

\item [DKG2: Knowledge Graph Aggregation.]
Design and implement mechanisms to encourage users to contribute their PKGs towards aggregated knowledge graphs, such as a concerted effort towards developing specific disease KGs. Blockchain-based incentive models that reward users for contributing to constructing such knowledge graphs, fostering collaborative efforts, and enriching the overall quality of shared knowledge are components of this goal. %[Models]

\item [DKG3: Knowledge Verification.]
Develop community-based and expert processes to verify the knowledge available in the global KGs. On the community front, it is critical to ensure that a knowledge item that was previously contributed through an individual has not been altered (either through error or with malicious intent), for instance via blockchain primitives, as explained in the previous section. %[Technologies]

\item [DKG4: Knowledge Validation.]
Validation of knowledge is paramount to ensure KG interoperability and the consumption of knowledge in target applications. By employing RDF and SHACL technologies, we ensure that the DKGs across different data stores conform to a specific template, thus, enabling their integration with community-supported KGs. %[Technologies]

\item [DKG5: Federated Querying.]
Explore and implement federated querying mechanisms, specifically utilising SPARQL, to enable efficient querying across integrated KGs. This process includes developing techniques to support various institutions, such as insurers, pharmaceutical companies, and medical research organisations, accessing and extracting insights from the knowledge graphs to enhance decision-making and advance their respective domains. %[Technologies]

\end{description}

%By pursuing these research goals, we aim to establish a Decentralised KG management framework that prioritises user privacy, fosters collaboration and knowledge sharing, incentivises participation, ensures robust verification and validation processes, and enables seamless querying for various institutions. 

\newpage
\noindent
\added{\textbf{Explainable Neuro-Symbolic AI.}}
\smallskip
%
%% We are now in page overflow territory.
%As illustrated earlier, hybrid AI systems that integrate symbolic reasoning and machine learning possess cognitive capabilities that can enhance transparency, explainability, and trustworthiness. However, despite the existing advancements, achieving trustworthy and accountable KG-based AI necessitates the development of novel methods that adapt hybrid systems to address current problem conditions while maintaining transparency in their integration. 
%This section enumerates the goals to be achieved in Explainable Neuro-Symbolic AI to contribute to the three pillars of KG-based AI for self-determination \autoref{fig:pillars}, i.e., Trust, Accountability, and Autonomy. These goals are defined based on the motivating use case \autoref{fig:scenario}, and they concert human-centric communication, results' accuracy, scalability, reproducibility, and generality.  \autoref{table:KG-AIAnalysis} depicts the results of the classification of these goals into short-, medium-, and long-term achievable goals. 
%
\begin{description}
\item[XNS1: User-dependent Recommendations.] Neuro-symbolic systems need to be empowered to transparently present results to the users according to their interests. For example, in our illustrative scenario, an individual may not expect the same level of detail in a health recommendation as a medical doctor or a community representative.
%Based on the development of explainable and neuro-symbolic systems, Accountability can be achieved in the short term. On the other hand, enabling traceability and transparency of the system's decisions for supporting Autonomy, and verification and validation for enhancing Trust can be completed in the middle period.  
\item [XNS2: Adaptive Hybrid AI.] Define models that can adaptively combine predictive models with logical reasoning, encompassing abilities such as generalisation and causal inference. For accountability, the neuro-symbolic system should explain when the combination of logical reasoning with a therapy bot or healthcare assistant will be beneficial. 
For autonomy, the neuro-symbolic system should include the user in the loop and consider their opinion in this decision. Finally, trust requires verifying and validating these decisions. 
\item [XNS3: Contextual-based Hybrid AI.] Equip neuro-symbolic systems with contextual knowledge, reasoning capabilities, and causal inference to effectively evaluate the strengths and limitations of machine learning components. This goal empowers the system to identify optimal combinations of statistical and symbolic AI methods, requiring the definition of causal models on top of KGs capable of combining reasoning over KGs with causal inference.
\item [XNS4: Symbolic Reasoning.] Employ inference processes, both inductive and deductive, on knowledge graphs to enable ML models, and LLMs in particular, to adjust hyper-parameters and a model's configuration, to new environments (i.e., Personal, community-based, and integrated healthcare KGs) and provide explanations for their decisions. Despite the advances of Automated Machine Learning (AutoML) systems (e.g., AutoML\footnote{\url{https://www.automl.org/}} and AutoWeka~\cite{KotthoffTHHL17}, to best of our knowledge, there are no developments for AutoML over KGs or for neuro-symbolic systems, which will enhance accountability, autonomy, and trust.
\item [\added{XNS5: Learning Transparency.}] \added{Investigate if existing XAI mechanisms can be tailored for learning transparency, such that it is possible to explain what action was take; how the decision making was performed; and why this was perceived as the outcome offering the greatest expected satisfaction.}
\end{description}

%\iffalse
\begin{table}[t] 
\caption{\textbf{Mapping of foundational requirements to pillars. A checkmark signifies that the corresponding requirement is necessary for answering a research question related to a pillar.} 
}
\label{table:mapping}
\centering
\footnotesize
\begin{tabular}{l|c|c|c|}
\cline{2-4} 

& \multicolumn{3}{c|}{\cellcolor[HTML]{7C98AA}\textbf{Machine-readable norms and policies}} 
\\ \cline{2-4} 

& {\cellcolor[HTML]{AED6F1} \textbf{Trust}} 
& {\cellcolor[HTML]{AED6F1}\textbf{Accountability}} 
& {\cellcolor[HTML]{AED6F1}\textbf{Autonomy}} 
\\ \hline

\multicolumn{1}{|l|}{\cellcolor[HTML]{FFFFC7}MRP1} & \multicolumn{1}{c|}{\cellcolor[HTML]{E8F7F8}\checkmark} & \multicolumn{1}{c|}{\cellcolor[HTML]{F2F3F6}} & \multicolumn{1}{c|}{\cellcolor[HTML]{FBF8EF}\checkmark}   
\\ \hline

\multicolumn{1}{|l|}{\cellcolor[HTML]{FFFFC7}MRP1} & \multicolumn{1}{c|}{\cellcolor[HTML]{E8F7F8}\checkmark} & \multicolumn{1}{c|}{\cellcolor[HTML]{F2F3F6}} & \multicolumn{1}{c|}{\cellcolor[HTML]{FBF8EF}\checkmark}  
\\ \hline

\multicolumn{1}{|l|}{\cellcolor[HTML]{FFFFC7}MRP3} & \multicolumn{1}{c|}{\cellcolor[HTML]{E8F7F8}} & \multicolumn{1}{c|}{\cellcolor[HTML]{F2F3F6}} & \multicolumn{1}{c|}{\cellcolor[HTML]{FBF8EF}\checkmark}  
\\ \hline

\multicolumn{1}{|l|}{\cellcolor[HTML]{FFFFC7}MRP4} & \multicolumn{1}{c|}{\cellcolor[HTML]{E8F7F8}\checkmark} & \multicolumn{1}{c|}{\cellcolor[HTML]{F2F3F6}} & \multicolumn{1}{c|}{\cellcolor[HTML]{FBF8EF}\checkmark}  
\\ \hline

\multicolumn{1}{|l|}{\cellcolor[HTML]{FFFFC7}MRP5} & \multicolumn{1}{c|}{\cellcolor[HTML]{E8F7F8}\checkmark} & \multicolumn{1}{c|}{\cellcolor[HTML]{F2F3F6}\checkmark} & \multicolumn{1}{c|}{\cellcolor[HTML]{FBF8EF}\checkmark}
\\ \hline

& \multicolumn{3}{c|}{\cellcolor[HTML]{7C98AA}\textbf{Decentralised Infrastructur}} 
\\ \cline{2-4} 

& {\cellcolor[HTML]{AED6F1} \textbf{Trust}} 
& {\cellcolor[HTML]{AED6F1}\textbf{Accountability}} 
& {\cellcolor[HTML]{AED6F1}\textbf{Autonomy}} 
\\ \hline

\multicolumn{1}{|l|}{\cellcolor[HTML]{FFFFC7}DI1} & \multicolumn{1}{c|}{\cellcolor[HTML]{E8F7F8}\checkmark} & \multicolumn{1}{c|}{\cellcolor[HTML]{F2F3F6}} & \multicolumn{1}{c|}{\cellcolor[HTML]{FBF8EF}\checkmark} 
\\ \hline

\multicolumn{1}{|l|}{\cellcolor[HTML]{FFFFC7}DI2} & \multicolumn{1}{c|}{\cellcolor[HTML]{E8F7F8}\checkmark} & \multicolumn{1}{c|}{\cellcolor[HTML]{F2F3F6}} & \multicolumn{1}{c|}{\cellcolor[HTML]{FBF8EF}\checkmark} 
\\ \hline

\multicolumn{1}{|l|}{\cellcolor[HTML]{FFFFC7}DI3} & \multicolumn{1}{c|}{\cellcolor[HTML]{E8F7F8}\checkmark} & \multicolumn{1}{c|}{\cellcolor[HTML]{F2F3F6}} & \multicolumn{1}{c|}{\cellcolor[HTML]{FBF8EF}\checkmark} 
\\ \hline

\multicolumn{1}{|l|}{\cellcolor[HTML]{FFFFC7}DI4} & \multicolumn{1}{c|}{\cellcolor[HTML]{E8F7F8}} & \multicolumn{1}{c|}{\cellcolor[HTML]{F2F3F6}\checkmark} & \multicolumn{1}{c|}{\cellcolor[HTML]{FBF8EF}} 
\\ \hline

\multicolumn{1}{|l|}{\cellcolor[HTML]{FFFFC7}DI5} & \multicolumn{1}{c|}{\cellcolor[HTML]{E8F7F8}} & \multicolumn{1}{c|}{\cellcolor[HTML]{F2F3F6}} & \multicolumn{1}{c|}{\cellcolor[HTML]{FBF8EF}\checkmark} 
\\ \hline

& \multicolumn{3}{c|}{\cellcolor[HTML]{7C98AA}\textbf{Decentralised KG Management.}} 
\\ \cline{2-4} 

& {\cellcolor[HTML]{AED6F1} \textbf{Trust}} 
& {\cellcolor[HTML]{AED6F1}\textbf{Accountability}} 
& {\cellcolor[HTML]{AED6F1}\textbf{Autonomy}} 
\\ \hline

\multicolumn{1}{|l|}{\cellcolor[HTML]{FFFFC7}DKG1} & \multicolumn{1}{c|}{\cellcolor[HTML]{E8F7F8}} & \multicolumn{1}{c|}{\cellcolor[HTML]{F2F3F6}} & \multicolumn{1}{c|}{\cellcolor[HTML]{FBF8EF}\checkmark} 
\\ \hline

\multicolumn{1}{|l|}{\cellcolor[HTML]{FFFFC7}DKG2} & \multicolumn{1}{c|}{\cellcolor[HTML]{E8F7F8}} & \multicolumn{1}{c|}{\cellcolor[HTML]{F2F3F6}\checkmark} & \multicolumn{1}{c|}{\cellcolor[HTML]{FBF8EF}\checkmark} 
\\ \hline

\multicolumn{1}{|l|}{\cellcolor[HTML]{FFFFC7}DKG3} & \multicolumn{1}{c|}{\cellcolor[HTML]{E8F7F8}\checkmark} & \multicolumn{1}{c|}{\cellcolor[HTML]{F2F3F6}} & \multicolumn{1}{c|}{\cellcolor[HTML]{FBF8EF}} 
\\ \hline

\multicolumn{1}{|l|}{\cellcolor[HTML]{FFFFC7}DKG4} & \multicolumn{1}{c|}{\cellcolor[HTML]{E8F7F8}\checkmark} & \multicolumn{1}{c|}{\cellcolor[HTML]{F2F3F6}} & \multicolumn{1}{c|}{\cellcolor[HTML]{FBF8EF}} 
\\ \hline

\multicolumn{1}{|l|}{\cellcolor[HTML]{FFFFC7}DKG5} & \multicolumn{1}{c|}{\cellcolor[HTML]{E8F7F8}} & \multicolumn{1}{c|}{\cellcolor[HTML]{F2F3F6}\checkmark} & \multicolumn{1}{c|}{\cellcolor[HTML]{FBF8EF}} 
\\ \hline

& \multicolumn{3}{c|}{\cellcolor[HTML]{7C98AA}\textbf{Explainable Neuro-Symbolic AI}} 
\\ \cline{2-4} 

& {\cellcolor[HTML]{AED6F1} \textbf{Trust}} 
& {\cellcolor[HTML]{AED6F1}\textbf{Accountability}} 
& {\cellcolor[HTML]{AED6F1}\textbf{Autonomy}} 
\\ \hline

\multicolumn{1}{|l|}{\cellcolor[HTML]{FFFFC7}XNS1} & \multicolumn{1}{c|}{\cellcolor[HTML]{E8F7F8}} & \multicolumn{1}{c|}{\cellcolor[HTML]{F2F3F6}\checkmark} & \multicolumn{1}{c|}{\cellcolor[HTML]{FBF8EF}} 
\\ \hline

\multicolumn{1}{|l|}{\cellcolor[HTML]{FFFFC7}XNS2} & \multicolumn{1}{c|}{\cellcolor[HTML]{E8F7F8}\checkmark} & \multicolumn{1}{c|}{\cellcolor[HTML]{F2F3F6}\checkmark} & \multicolumn{1}{c|}{\cellcolor[HTML]{FBF8EF}} 
\\ \hline

\multicolumn{1}{|l|}{\cellcolor[HTML]{FFFFC7}XNS3} & \multicolumn{1}{c|}{\cellcolor[HTML]{E8F7F8}\checkmark} & \multicolumn{1}{c|}{\cellcolor[HTML]{F2F3F6}\checkmark} & \multicolumn{1}{c|}{\cellcolor[HTML]{FBF8EF}} 
\\ \hline

\multicolumn{1}{|l|}{\cellcolor[HTML]{FFFFC7}XNS4} & \multicolumn{1}{c|}{\cellcolor[HTML]{E8F7F8}} & \multicolumn{1}{c|}{\cellcolor[HTML]{F2F3F6}\checkmark} & \multicolumn{1}{c|}{\cellcolor[HTML]{FBF8EF}} 
\\ \hline

\multicolumn{1}{|l|}{\cellcolor[HTML]{FFFFC7}XNS5} & \multicolumn{1}{c|}{\cellcolor[HTML]{E8F7F8}} & \multicolumn{1}{c|}{\cellcolor[HTML]{F2F3F6}\checkmark} & \multicolumn{1}{c|}{\cellcolor[HTML]{FBF8EF}} 
\\ \hline

\end{tabular}
\end{table}
%\fi

\subsection{\added{AI for Self-determination}}

\added{The identified foundational research topic challenges and opportunities can be used to better contextualise concrete goals in relation to trust, accountability, and autonomy from a KG-based AI for self-determination perspective. An overview of this mapping, which is depicted in Table~\ref{table:mapping}, is provided by attempting to answer the overarching questions that guide our vision paper.}

\begin{description}    
    \item[\added{(Q1) What are the key requirements for an AI system to produce trustable results?}] \added{From a trust perspective, it is important that machine-readable policies faithfully represent the human policies (MRP1) in a manner that can be verified automatically (MRP2). Regardless of whether systems are automated or semi-automated, we need to be able to verify that processes behave as expected (MRP4) and any misuse can be detected and rectified (MRP5). Trust could potentially be facilitated via auditing (DI1) and tracing (DI2), as well as certification mechanisms that support decency checks (DI3) and (semi-)automated knowledge verification (DKG3) and validation (DKG4) techniques. While human involvement is paramount to establishing trust in adaptive (XNS2) and contextualised (XNS3) hybrid AI.}   
    \item[\added{(Q2) How can AI be made accountable for its decision-making?}] \added{The first step to achieving accountability is to ensure it is possible to detect if any party violated policies and norms (MRP5) and that the recommendations given and decisions taken using both induction and deduction (XNS4) are comprehensible from a users perspective, for instance via user focuses recommendations (XNS1), providing explanations for recommendations and decisions (XNS2), facilitating learning transparency (XNS5), and contextualisation based on causal inference (XNS3). Considering that machines can only work with the knowledge that it has at hand, it is important that systems are able to integrate knowledge from disparate sources (DI4), and are capable of querying (DKG5) and aggregating (DKG2) relevant sources.}   
    \item[\added{(Q3) How can citizens maintain autonomy as users or subjects of KG-based AI systems?}] \added{Citizens' autonomy in a KG-based AI context is necessary to ensure that humans are able to control not only who has access to their personal data, but also that its usage is in line with existing regulatory requirements. The could be achieved with automated compliance checking (MRP4) and misuse detection (MRP5) built on top of machine-readable policies (MRP1) and evaluation mechanisms (MRP2). Negotiation could potentially enable organisations to gain access to better quality data (MRP3) or to foster collaboration via aggregation (DKG2) and strong privacy guarantees via anonymisation (DKG1). While, self-sovereign identities (DI5), auditing (DI1), tracing (DI2), and decency certification (DI3) have a major role to play when it comes to continuous monitoring.}
\end{description}

%\sabrina{Here we can revisit the questions from the introduction. Given that it is not possible to answer these questions I suggest that we could simply map the agenda items in 5.1 to the research questions and add some supporting text around the need for models, technologies, and standards.}

\section{Conclusion}
\label{sec:conclusion}
This paper presents a compelling argument for integrating KG-based AI to empower individuals' self-determination and benefit society. This overarching goal is supported by three fundamental pillars: trust, accountability, and autonomy. We advocate the foundations of  these pillars require focused research in four areas: machine-readable norms and policies, decentralised infrastructure, decentralised KG management, and explainable neuro-symbolic AI. By drawing on a concrete scenario within the healthcare domain, we demonstrate the relevance of each foundational topic and outline a comprehensive research agenda for each of them. 

We aspire for the insights presented in this paper to catalyse the creation of AI services that genuinely support citizens while upholding their rights. Responsible advancement of the foundational topics is crucial to ensure that future KG-based AI solutions are comprehensive and possess the qualities of being traceable, verifiable, and interpretable. It is essential that relevant legislation, such as the EU AI Act, provides clear guidance to steer the development of these forthcoming applications, emphasising the need for accurate, reliable, and transparent AI systems.
Within this context, we recognise the Semantic Web community as uniquely positioned to drive transformative change and contribute solutions illuminating opaque AI models' workings. Through this concerted effort, we envision a paradigm shift in KG management and analytics that establishes KG-based AI to empower individuals in their pursuit of self-determination.

%In this paper, we motivate the need for KG-Based AI to be a force for individuals' self-determination and the benefit of society. We posit this goal is built upon three pillars: trust, accountability and autonomy, that in turn are achieved by researching on four foundational topics: Machine Readable norms and policies, Decentralised Infrastructure, Decentralised KG Management and Explainable Neuro-Symbolic AI. Grounding on an illustrative scenario in the healthcare domain, we describe how each foundational topic related to the pillars and propose a research agenda for each of them.  

\subparagraph*{Acknowledgements}

Luis-Daniel Ib\'{a}\~nez is partially funded by the European Union's Horizon Research and Innovation Actions under Grant Agreement nº 101093216 (UPCAST).  Sabrina Kirrane is funded by the FWF Austrian Science Fund and the Internet Foundation Austria under the FWF Elise Richter and netidee SCIENCE programmes as project number V 759-N. Oshani Seneviratne is partially funded by NSF IUCRC CRAFT center research grant (CRAFT Grant \#22008) and the Algorand Centres of Excellence programme managed by Algorand Foundation. The opinions expressed in this publication do not necessarily represent the views of NSF IUCRC CRAFT or the Algorand Foundation.  Maria-Esther Vidal is partially funded by Leibniz Association, program ``Leibniz Best Minds: Programme for Women Professors'', project TrustKG-Transforming Data in Trustable Insights; Grant P99/2020.

%%
%% Bibliography
%%
%% Please include DOIs when available
%%  as well as unabbreviated first names.

%% Either use bibtex (recommended)
\bibliographystyle{plainurl}
\bibliography{main}

%% .. or use bibitems explicitly
%%
%%\begin{thebibliography}{1}
%%
%%\bibitem{Leclerc22}
%%Guillaume Leclerc.
%%\newblock {Graphs for Data}.
%%\newblock In Amélie Jeunet and Tom Finn, editors, {\em {ICBE} `22: The 29th
%%  International Conference on Bibliographical Examples, France, October 29-30,
%%  2022}, pages 2233--2244. {ACM}, 2022.
%%\newblock \href {https://doi.org/10.1145/9876543.2109876}
%%  {\path{doi:10.1145/9876543.2109876}}.
%%
%%\bibitem{MurphyL20}
%%Niamh Murphy and Seamus Lynch.
%%\newblock {Graphs for Knowledge}.
%%\newblock {\em Journal of Bibliographical Examples}, 10(5):872--903, 2020.
%%\newblock \href {https://doi.org/10.1007/s12345-678-00009-x}
%%  {\path{doi:10.1007/s12345-678-00009-x}}.
%%
%\end{thebibliography}

\end{document}